%% file: iclr2026_conference.tex
\documentclass{article} 
\usepackage[utf8]{inputenc}
\usepackage{textgreek}
\usepackage{iclr2026_conference,times}

\input{math_commands.tex}

\usepackage{hyperref}
\usepackage{url}

\usepackage{textcomp}
\usepackage{subcaption}
\usepackage{adjustbox}
\usepackage{pifont}
\usepackage{wrapfig}
\usepackage[table,xcdraw]{xcolor}
\usepackage{float}
\usepackage{dsfont}
\usepackage{wrapfig}
\usepackage{paralist}
\definecolor{my_green}{HTML}{2a9d8f}
\definecolor{my_red}{HTML}{e76f51}
\definecolor{my_yellow}{HTML}{e9c46a} 

\usepackage{listings}
\usepackage{xcolor}

\lstdefinelanguage{json}{
    basicstyle=\ttfamily\small,
    numbers=none,
    breaklines=true,
    frame=single,
    backgroundcolor=\color{gray!5},
    literate=
     *{0}{{{\color{black}0}}}{1}
      {1}{{{\color{black}1}}}{1}
      {2}{{{\color{black}2}}}{1}
      {3}{{{\color{black}3}}}{1}
      {4}{{{\color{black}4}}}{1}
      {5}{{{\color{black}5}}}{1}
      {6}{{{\color{black}6}}}{1}
      {7}{{{\color{black}7}}}{1}
      {8}{{{\color{black}8}}}{1}
      {9}{{{\color{black}9}}}{1}
      {:}{{{\color{black}{:}}}}{1}
      {,}{{{\color{black}{,}}}}{1}
      {"}{{{\color{orange}{"}}}}{1},
    morestring=[b]",
    stringstyle=\color{blue},
    showstringspaces=false
}

\usepackage{enumitem}
\usepackage{booktabs}
\usepackage{mathtools}
\usepackage{multirow}
\usepackage[ruled,vlined]{algorithm2e}
\usepackage{multicol}

\usepackage[most]{tcolorbox}
\usepackage{float} 
\newcounter{mybox}
\renewcommand{\themybox}{\arabic{mybox}} 
\counterwithout{mybox}{section}

\newtcolorbox[use counter=mybox]{mytbox}[2][]{%
  colback=gray!5!white, colframe=gray!75!black,
  title={Box~\themybox.~#2}, #1, breakable, 
  float, floatplacement=t,
}

\title{Endowing GPT-4 with a Humanoid Body: \\Building the Bridge Between \\Off-the-Shelf VLMs and the Physical World}

\author{Yingzhao Jian, Zhongan Wang, Yi Yang, Hehe Fan
\thanks{Corresponding author: \texttt{hehefan@zju.edu.cn}}\\
College of Computer Science and Technology\\
Zhejiang University\\
Hangzhou, Zhejiang, China \\
\texttt{hehefan@zju.edu.cn} \\
}

\iclrfinalcopy 
\renewcommand{\iclrfinalcopy}{}
\makeatletter
\renewcommand{\@maketitle}{
    \vspace*{-2cm}
    \begin{center}
        {\LARGE \bf \@title \par}
        \vskip 0.5em
        {\large
            \lineskip .5em
            \begin{tabular}[t]{c}
                \@author
            \end{tabular}\par}
        \vskip 1em
    \end{center}
    \par
    \vskip 1.5em
}
\makeatother
\begin{document}

\maketitle

\input{sections/0_abstract}

\input{sections/1_introduction}
\input{sections/2_related_work}
\input{sections/3_method}
\input{sections/4_experiment}
\input{sections/5_conclusion}

\bibliography{iclr2026_conference}
\bibliographystyle{iclr2026_conference}

\newpage
\clearpage

\appendix
\input{sections/A_sup_related_work}
\input{sections/B_datasets}
\input{sections/C_compiler}
\input{sections/D_executor}
\input{sections/E_com_methods}
\input{sections/F_sup_experiment}

\end{document}

%% file: math_commands.tex
\usepackage{amsmath,amsfonts,bm}

\def\eqref#1{equation~\ref{#1}}

\def\1{\bm{1}}

\def\vl{{\bm{l}}}

\def\vp{{\bm{p}}}

\def\vs{{\bm{s}}}

\def\mM{{\bm{M}}}

\def\mO{{\bm{O}}}

\def\mS{{\bm{S}}}

\DeclareMathAlphabet{\mathsfit}{\encodingdefault}{\sfdefault}{m}{sl}
\SetMathAlphabet{\mathsfit}{bold}{\encodingdefault}{\sfdefault}{bx}{n}

\def\sJ{{\mathbb{J}}}

\def\sR{{\mathbb{R}}}

\newcommand{\E}{\mathbb{E}}



%% file: sections/0_abstract.tex
\begin{abstract}
Humanoid agents often struggle to handle flexible and diverse interactions in open environments.
A common solution is to collect massive datasets to train a highly capable model, but this approach can be prohibitively expensive.
In this paper, we explore an alternative solution: empowering off-the-shelf Vision-Language Models (VLMs, such as GPT-4) to control humanoid agents, thereby leveraging their strong open-world generalization to mitigate the need for extensive data collection. 
To this end, we present \textbf{BiBo} (\textbf{B}uilding humano\textbf{I}d agent \textbf{B}y \textbf{O}ff-the-shelf VLMs). It consists of two key components: (1) an \textbf{embodied instruction compiler}, which enables the VLM to perceive the environment and precisely translate high-level user instructions (e.g., {\small\itshape ``have a rest''}) into low-level primitive commands with control parameters (e.g., {\small\itshape ``sit casually, location: (1, 2), facing: 90$^\circ$''}); and (2) a diffusion-based \textbf{motion executor}, which generates human-like motions from these commands, while dynamically adapting to physical feedback from the environment.
In this way, BiBo is capable of handling not only basic interactions but also diverse and complex motions. 
Experiments demonstrate that BiBo achieves an interaction task success rate of 90.2\% in open environments, and improves the precision of text-guided motion execution by 16.3\% over prior methods. The code will be made publicly available. 
\end{abstract}

%% file: sections/1_introduction.tex
\section{Introduction}
\label{sec:intro}

Humanoid agents, as a medium between digital intelligence and the physical world, have attracted extensive research interest, particularly in the domains of scene perception~\citep{huang2024embodied, qi2025vln} and interaction~\citep{xiao2023unified,tevet2024closd}. With recent advances, humanoid agents are capable of performing text-guided motions~\citep{tevet2022human, yuan2024mogents} and executing interactive tasks under predefined plans~\citep{xu2024humanvla,pan2025tokenhsi}. However, flexibly handling user-intended scene interactions in open and dynamic physical environments remains a significant challenge. A straightforward strategy is to collect large-scale human–scene interaction data~\citep{bhatnagar2022behave,jiang2024scaling} and train a highly capable model, as commonly done for robotic arms or wheeled platforms~\citep{firoozi2025foundation,team2025robobrain}. Unfortunately, due to the structural complexity of humanoid bodies and the vast diversity of physical world, such data-centric scaling becomes prohibitively expensive and difficult to generalize.

In contrast, off-the-shelf general-purpose Vision–Language Models (VLMs), such as GPT-4~\citep{achiam2023gpt}, Gemini~\citep{team2023gemini}, and Qwen~\citep{bai2023qwen}, demonstrate open-world reasoning and adaptability across a wide variety of tasks,
without specific finetuning. This observation naturally raises an intriguing question: \textit{Can we bypass costly data collection by directly leveraging these powerful off-the-shelf VLMs to control humanoid agents, thereby enabling more versatile  interaction in the physical world?}

\begin{figure}[t!]
\vspace{-3mm}
    \centering
    \includegraphics[width=1.0\textwidth]{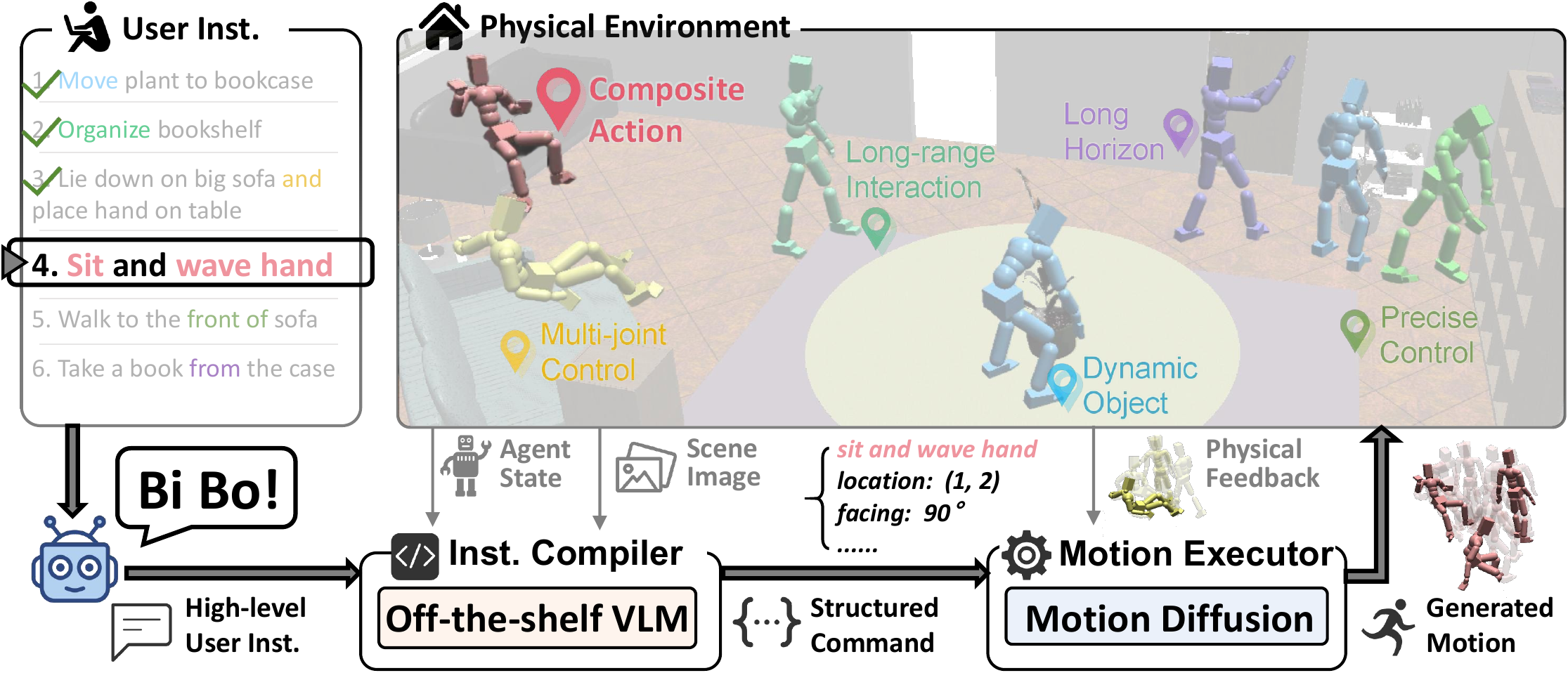}
    \caption{\textbf{BiBo} is a humanoid agent powered by an off-the-shelf VLM. It consists of an embodied instruction compiler (Inst. Compiler) and a diffusion-based motion executor. When the user provides a high-level instruction, the compiler observes the environment and translates it into the structured command for the executor. The executor then generates future motions for the humanoid agent, conditioned on both the command and the physical feedback from the environment. In this way, BiBo is able to perform diverse types of physical scene interactions.}
    \label{fig:fig1}
    \vspace{-5mm}
\end{figure}

Motivated by this question, we introduce BiBo (\textbf{B}uilding humano\textbf{I}d agent \textbf{B}y \textbf{O}ff-the-shelf VLMs), a framework designed to endow off-the-shelf VLMs with the capability to control humanoid agents. BiBo is composed of two core components: 1) an \textbf{VLM-driven embodied instruction compiler} and 2) a \textbf{diffusion-based motion executor}, which jointly bridge the gap between high-level human instructions and low-level motor control required for physical humanoid interactions. This design is conceptually analogous to a computer, where the compiler and the assembler work together, operating the hardware to accomplish the tasks specified by programming language.

In computing system, a compiler translates source code written by high-level programming languages into low-level assembled language. Inspired by this, as in Fig.~\ref{fig:fig1}, the embodied instruction compiler in BiBo is designed to translate high-level natural language instructions into low-level executor commands, according to environmental context.
To achieve this, the compiler first represents an action as a structured set of descriptors, encompassing the motion caption, key joint configurations, and other contextual details. Building on this structured representation, the compiler drives the VLM to reason over each descriptor in a coarse-to-fine manner, thereby producing an accurate and structured command that specifies the user intented action.

This generated command is then passed to the motion executor, which functions much like an assembler.
Just as translating assembly commands into machine code, the motion executor interprets commands into full-body humanoid motions. Unlike a rigid rule-based assembler, our executor leverages a diffusion generator. Each time it receives a command, the generator extends future joint trajectories from the current motion, enabling diverse motion style and on-the-fly control.

However, during execution, collisions or external forces may cause the actual executed motion to deviate from the initially generated sequence. To handle such feedback, prior approaches~\citep{tevet2024closd,chen2024taming} extending future joint trajectories from the executed motion, rather than from those previously generated but unexecuted. This strategy enforces the diffusion to account for environmental context, but also introduces discontinuities between the current and previous generated motions.
We resolve this by developing a novel application of the Latent Diffusion Model (LDM)~\citep{chen2023executing}. In our method, the diffusion extends future latent from the actual executed motion, enabling environmental awareness. A Variational Autoencoder (VAE) jointly decodes the latents of the previous and current generated motions, ensuring smooth transition. 

According to the experiments, BiBo achieves an interaction task success rate of 90.2\% under random generated physical environments using an off-the-shelf VLM (i.e., GPT-4o). Moreover, BiBo improves the precision of text-guided motion execution by 16.3\% over prior methods. It handles complex motion execution, while also enabling infinite long-sequence synthesis and real-time interactive control through user instructions. In summary, our main contributions are threefold:

\begin{itemize}
    \item We empower off-the-shelf VLMs for humanoid control through an embodied instruction compiler and a diffusion-based motion executor, bridging the gap between general-purpose VLM reasoning and low-level physical execution.  
\end{itemize}
\begin{itemize}
    \item The compiler introduces a structured humanoid action representation, enabling coarse-to-fine embodied reasoning, while advancing humanoid behavior planning and modeling.  
    \item We develop a novel application of LDM to incorporate environmental feedback in motion generation, achieving state-of-the-art unlimited-length motion synthesis and offering insights for applying LDMs in broader domains.  
\end{itemize}

%% file: sections/2_related_work.tex
\section{Related Work}

\subsection{Human Scene Interaction} 
When interacting with scene, humanoids perceive environments through training on real-world data, reinforcement learning (RL), and large language models (LLMs). 
Recent advances combine them: (1) using LLM to guide RL policies~\citep{xiao2023unified,shi2024large}, but still limits motion diversity; (2) using RL trackers~\citep{luo2022universal, luo2023perpetual} to follow diffusion-generated motions~\citep{tevet2024closd}, but causes discontinuities between generated and tracked motions. BiBo employs an off-the-shelf VLM to guide a latent diffusion model (LDM), promoting generalization and diversity, which achieving both smoothness and physical plausibility.

\subsection{Text to Motion Generation}

In text-to-motion, approaches can be broadly categorized into fixed- and arbitrary-length generation. For fixed-length generation, frameworks such as VAEs~\citep{petrovich21actor, bie2022hit}, masked modeling~\citep{pinyoanuntapong2024mmm,guo2024momask}, and diffusion models~\citep{tevet2022human,zhang2024motiondiffuse,dai2024motionlcm} have been extensively explored. However, humanoids perform continuous arbitrary-length motion following user commands. To this end, some works adopt autoregressive next-token prediction~\citep{jiang2023motiongpt,zhang2023generating}, achieving high fidelity, while others~\citep{chen2024taming,han2024amd,xiao2025motionstreamer} employ diffusion to extend future joint trajectories from past motion, improving efficiency. BiBo use latent diffusion with few denoising steps, enabling both high-fidelity generation and real-time control.

%% file: sections/3_method.tex
\section{Method}
\subsection{Overview}

BiBo is a humanoid agent powered by an off-the-shelf Vision-Language Model (VLM). As shown in Fig.~\ref{fig:fig1}, it comprises two main components: an embodied instruction compiler and a diffusion-based motion executor.
Given a high-level user instruction, the compiler first collects observations of the current scene, and then prompts the VLM to generate a caption of the next motion to be executed. Next, the compiler guides the VLM to refine motion details through a three-stage visual question-answering (VQA) process. Finally, it formats these details into a command based on a structured motion representation, thereby instructing the executor to generate the corresponding motion.

The executor takes the command as a condition, extending future joint trajectories from the current motion. Due to collisions or external forces, the actual executed motion may differ from the predicted trajectories. To adapt to this feedback, we incorporate the actual performed motion into diffusion by developing a novel application of the Latent Diffusion Model (LDM). In diffusion, the model extends future motion latents from the actual executed motion, thereby adapting to the scene feedback; in the VAE, the model jointly decodes the previously generated and currently executed motions, ensuring smooth transitions between previous and current generated motions.

\begin{figure}[t!]
    \centering
    \vspace{-4mm}
    \includegraphics[width=1.0\textwidth]{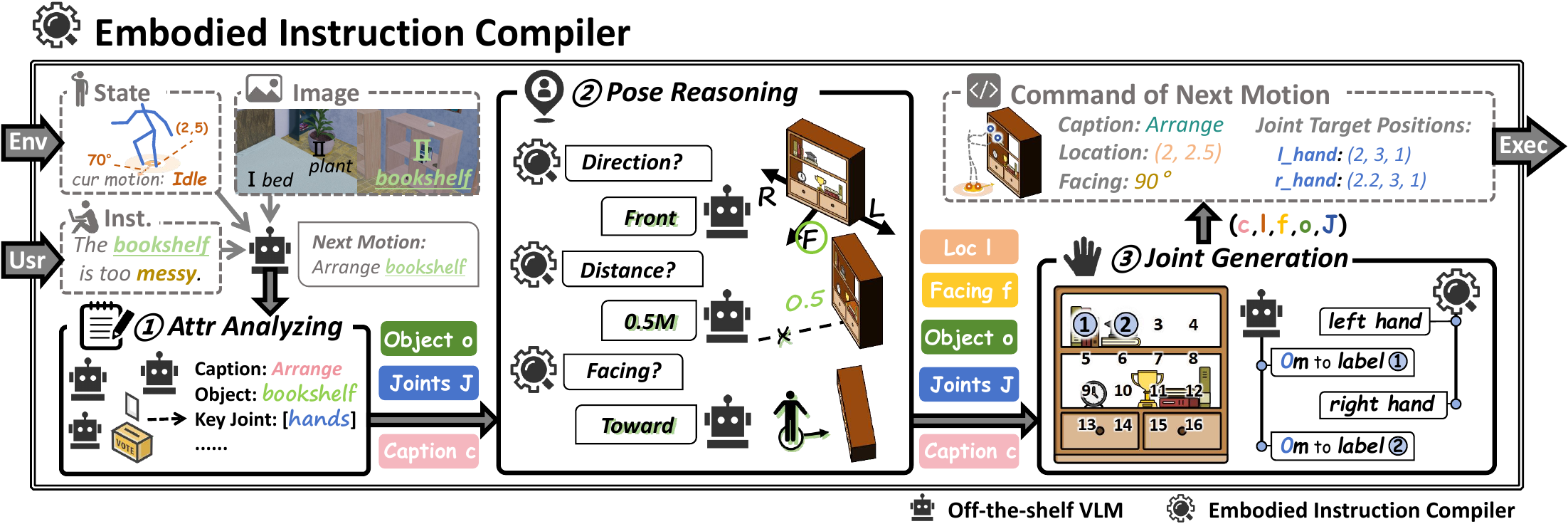} 
    \caption{The \textbf{embodied instruction compiler} takes in user instructions and environmental observations, and directs the VLM to generate the next motion command through a structured three-stage visual question–answering process. In the first stage, it analyzes the basic attributes of the motion (e.g., caption, key joints, target object). In the second stage, it reasons about the agent’s pose during the interaction. Finally, it specifies the target positions for the key joints. }       
    \label{fig:fig2} 
    \vspace{-4mm}
\end{figure}

\subsection{Embodied Instruction Compiler}
\label{sec:met:com}
As in Fig.~\ref{fig:fig2}, the embodied instruction compiler enables the VLM to translate high-level user instructions into low-level executor commands, based on environmental observation. It consists of a three-stage visual question–answering process. The VLM first determines the next motion to execute, and analyzes its basic attributes, such as the motion caption and target object. Then, it reasons about the agent pose. Finally, the VLM locates the target positions of the key joints.
The output of the compiler is a executor command $\mathcal{C}$, including the caption $c$, location $\vl \in \sR^{2}$, facing direction $f \in [-\pi, \pi]$, and joint targets $\sJ = \{(j, \vp_j): j~ \mathrm{is\ a\ key\ joint}\}$, where $\vp_j \in \sR^3$ is the joint target:

\vspace{-4mm}
\begin{equation}
\mathcal{C} = \{c,~\vl,~f,~\sJ\}.
\end{equation}
\vspace{-4mm}

The command $\mathcal{C}$ serves as a structured and simplified humanoid action representation, which reduces generation complexity while preserving the key information of a motion, controlling the diffusion to generate an interactive motion that fulfills the instruction. Details can be found in Sec.~\ref{sec:emb_ins_com}.

\textbf{Basic Attribute Analysis.} 
In this step, the compiler inputs the user instruction with scene images and agent’s status, and prompts the VLM to analyze the attributes of the next motion to be executed. These attributes include a motion caption $c$, an anchor object $o$ for agent localization, the key joints $j$ involved, and other complementary details that facilitate subsequent reasoning. To enhance accuracy, the final result is selected through majority voting across five parallel VLM instances.

\textbf{Agent Pose Reasoning.} 
The pose refers to the location $\vl$ and facing direction $f$ of the agent. To predict the pose, 
directly outputting coordinates and angles is a straightforward way. However, current off-the-shelf VLMs struggle to handle numbers like 3D coordinates~\citep{huang2024chat, qi2025gpt4scene}.
As a result, we transform it into a visual identification task, which is more familiar to VLMs. As in Fig.~\ref{fig:fig2}, we put labels around the anchor object $o$, each corresponds to a position or direction. By choosing a label from the image, the VLM roughly locates the agent in the scene.

\textbf{Key Joint Generation.}
When interacting with objects, we typically focus on a few key joints and their relative position to specific target points (e.g., when using a hand dryer, the hands are placed about $0.2$m beneath the air outlet). Inspired by this, for each key joint, we first place a $8\times8$ grid of labels over the image of the anchor object, and allow the VLM to select one as the target point. Next, the VLM generates the joint's direction and distance relative to the target point. We provide a set of predefined directions: {\small\itshape[up, down, left, right, forward, backward, toward the object center, along the surface normal]}. This simplification reduces generation difficulty while covering most cases.

\subsection{Motion Diffusion Executor}
\label{sec:met:mde}
The Motion Diffusion Executor is a Latent Diffusion Model, composed of a VAE and a Diffusion module. When command $\mathcal{C}$ comes, the VAE first encodes both the previously generated motion $\mM_g \in \sR^{F \times D}$ and the actual executed motion $\mM_a \in \sR^{F \times D}$ (i.e. the execution result of $\mM_g$ under physical environment) into latent representations $\mS_g \in \sR^{L\times H}$ and $\mS_a \in \sR^{L\times H}$, respectively:

\vspace{-4mm}
\begin{equation}
\mS_a = \text{Encoder}(\mM_a), \quad \mS_g = \text{Encoder}(\mM_g).
\end{equation}
\vspace{-5mm}

$F$ is the number of frames of the motion. $L$ is the number of latent tokens, where each token correspond to $\frac{F}{L}$ frames. $D$ and $H$ are the dimension of motion frame and latent. Then, the command $\mathcal{C}$, together with the latent of the executed motion $\mS_a$ guide the denoising process to produce the latent of future motion $\mS_f \in \sR^{L\times H}$. The latents of the previous and current generated motion $\mS_g$ and $\mS_f$ are jointly decoded by the VAE to obtain the future joint trajectories $\mM_f\in \sR^{F \times D}$:

\begin{figure}[t!]
\vspace{-3mm}
    \centering
    \includegraphics[width=1.0\textwidth]{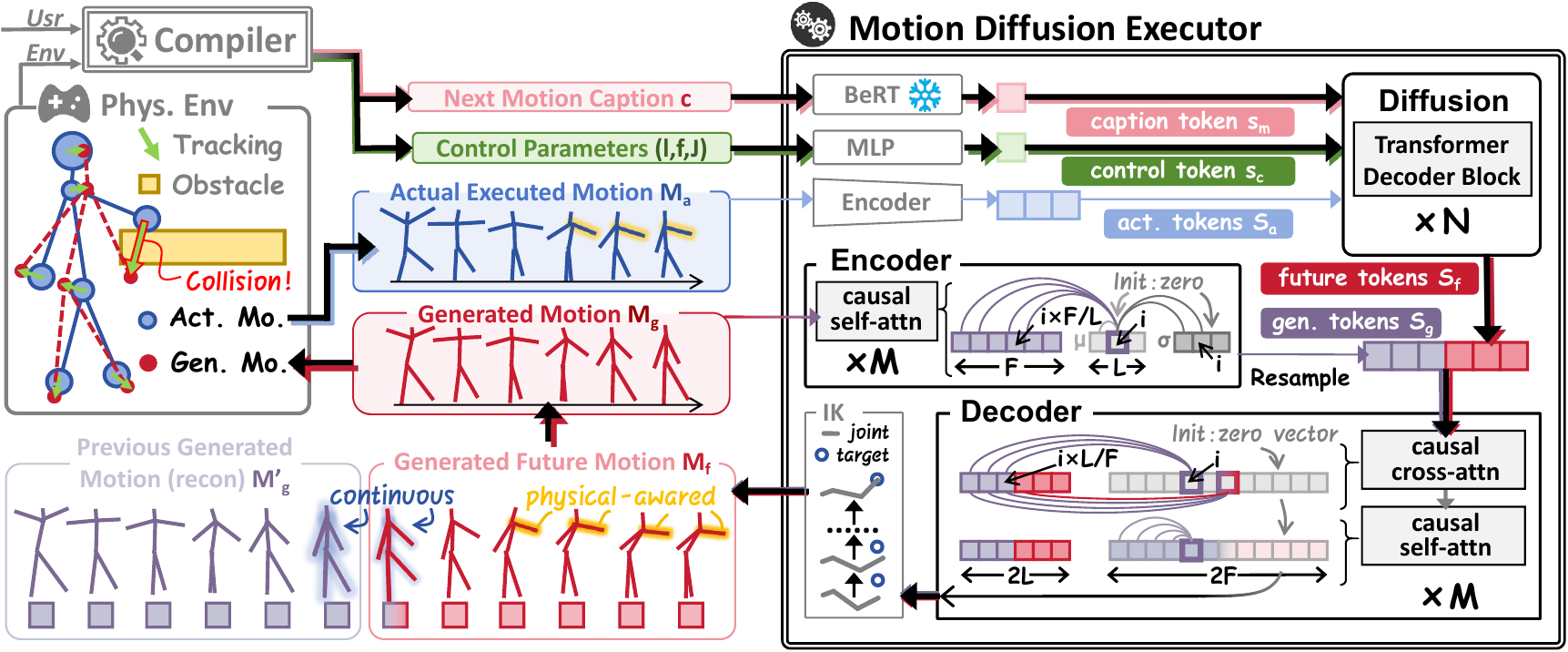} 
    \caption{The \textbf{motion executor} is a Latent Diffusion Model. When receiving the command (motion caption and control parameters) from the compiler, the Diffusion extends the future latents $\mS_f$ from the actual executed motion tokens $\mS_a$, conditioned on the command tokens $\vs_m$ and $\vs_c$. Then, the previous and newly generated latents are jointly decoded by the VAE decoder. The decoder use casual attention, where each motion frame or latent token can only attend to its preceding tokens or frames. After IK optimization, a tracking policy drive humanoid joints to execute the newly generated motion $\mM_f$ in physical environment, producing the next execution result.}     
    \label{fig:fig3}
    \vspace{-8mm}
\end{figure}

\vspace{-4mm}
\begin{equation}
\mS_f = \text{Diffusion}(\mathcal{C}, \mS_a), 
\quad 
[\mM^{'}_g: \mM_f] = \text{Decoder}([\mS_g:\mS_f]).
\label{eq:diffusion-decoder}
\end{equation}
\vspace{-5mm}

$\mS_f \in \sR^{L\times H}$ is the generated latent of future motion. $\mM^{'}_g \approx \mM_g \in \sR^{F \times D}$ is the reconstructed previous generated motion. [:] represents the concatenation across length (or frame) dimension. 
By conditioning on $\mS_a$, the future motion $\mM_f$ is guided to account for physical feedback in $\mM_a$, while joint decoding enforces its continuity with $\mM_g$.

We additionally use inverse kinematics (IK) post-optimization to improve control accuracy. Finally, a reinforcement learned tracking policy drives humanoid joints to follow the generated joint trajectories in physical scene. For more details, please refer to Sec.~\ref{sec:exe:ik}.

\textbf{VAE Design.} We use Transformer encoder-decoder architecture and propose causal self-cross attention. Specifically, during the attention process, each latent token or motion frame can only attend to its preceding tokens or frames. This design ensures the continuity between the previous and current generated motion $\mM_g$ and $\mM_f$. Specifically, as in Eq.~\ref{eq:diffusion-decoder}, since the VAE ensures continuity in its decoded motions, the the generated future motion $\mM_f$ is continuous with reconstructed previous motion $\mM^{'}_g$.  Moreover, due to the causal mechanism:

\vspace{-3mm}
\begin{equation}
[\mM^{'}_g:\mM_f] = \text{Decoder}([\mS_g:\mS_f]) 
~~\Rightarrow~~ 
\mM^{'}_g = \text{Decoder}(\mS_g)~~\Rightarrow~~ \mM^{'}_g\approx\mM_g.
\end{equation}
\vspace{-5mm}

$\mM_f$ is continuous with $\mM^{'}_g$, and $\mM^{'}_g \approx \mM_g$. Therefore, $\mM_f$ is also continuous with $\mM_g$.

%% file: sections/4_experiment.tex
\section{Experiments}

In this section, we analyze the capabilities of BiBo from two perspectives: task completion and motion quality. For task completion, we establish a set of tasks, and assess the success rate of BiBo and comparison methods under random generated scenes. For motion quality, we adopt standard motion quality metrics to quantitatively evaluate the synthesized and executed motions, and conduct case studies with visualizations to analyze the qualitative results.

\subsection{Task Completion}
\label{sec:task_completion}
\textbf{Task Setting.} We define six types of single interaction tasks. Task success if the required criterion is met for over 30 frames. The tasks include:

\textbullet~\textit{Reach} is considered successful if the agent reaches within 0.5 meters of the target location.

\begin{table}
\centering
\caption{Comparison of \textbf{task success rates} for different methods under randomly generated scenes and initial poses. A single task involves navigating to the interaction position and performing the interaction, whereas a composite task consists of multiple simultaneous or sequential single interactions. BiBo (our) performs online planning during evaluation, while other methods use ground truth action plan. The \textbf{bold} and \underline{underline} represent the best and second-best performance, respectively. BiBo achieves the highest success rate across all tasks.}
\scriptsize
\resizebox{1.0\textwidth}{!}{
\begin{tabular}{l|cccccc|ccc}
\hline
\multirow{2}{*}{\textbf{Method (\%)}} & \multicolumn{6}{c|}{\textbf{Single Interaction}} & \multicolumn{3}{c}{\textbf{Composite Task}} \\
\cline{2-7} \cline{8-10}
 & Reach $\uparrow$ & Watch $\uparrow$ & Sit $\uparrow$ & Sleep $\uparrow$ & Touch $\uparrow$ & Lift $\uparrow$ & Simple $\uparrow$& Medium $\uparrow$ & Hard $\uparrow$ \\
\hline
UniHSI\citep{xiao2023unified}         & 93.28 &   -   & 81.03 & 85.11 & 69.62 &   -   &   -   &   -   &   -   \\
HumanVLA  \citep{xu2024humanvla}     & 56.58 &   -   &   -   &   -   &   -   & 44.90 &   -   &   -   &   -   \\
TokenHSI   \citep{pan2025tokenhsi}    & 94.55 &   -   & 72.95 & 33.33 &   -   & 48.19 &   -   &   -   &   -   \\
CLoSD     \citep{tevet2024closd}     & 85.83 & 87.76 & 76.99 & 34.67 & 42.55 &  7.71 & 26.47 &  7.05 &  2.38 \\
\hline
BiBo (ours)         & \textbf{99.18} & \textbf{99.62} & \underline{95.84} & \textbf{94.89} & \underline{86.05} & \underline{65.42} & \underline{58.82} & \underline{36.54} & \underline{27.78} \\
BiBo (ours, GT plan) & \underline{98.91} & \underline{99.06} & \textbf{96.75} & \underline{93.33} & \textbf{87.23} & \textbf{70.41} & \textbf{61.76} & \textbf{44.23} & \textbf{42.86} \\
\hline
\end{tabular}
}
\label{tab:tab1}
\vspace{-8mm}
\end{table}

\begin{wrapfigure}{r}{0.4\textwidth}
  \centering
  \includegraphics[width=0.4\textwidth]{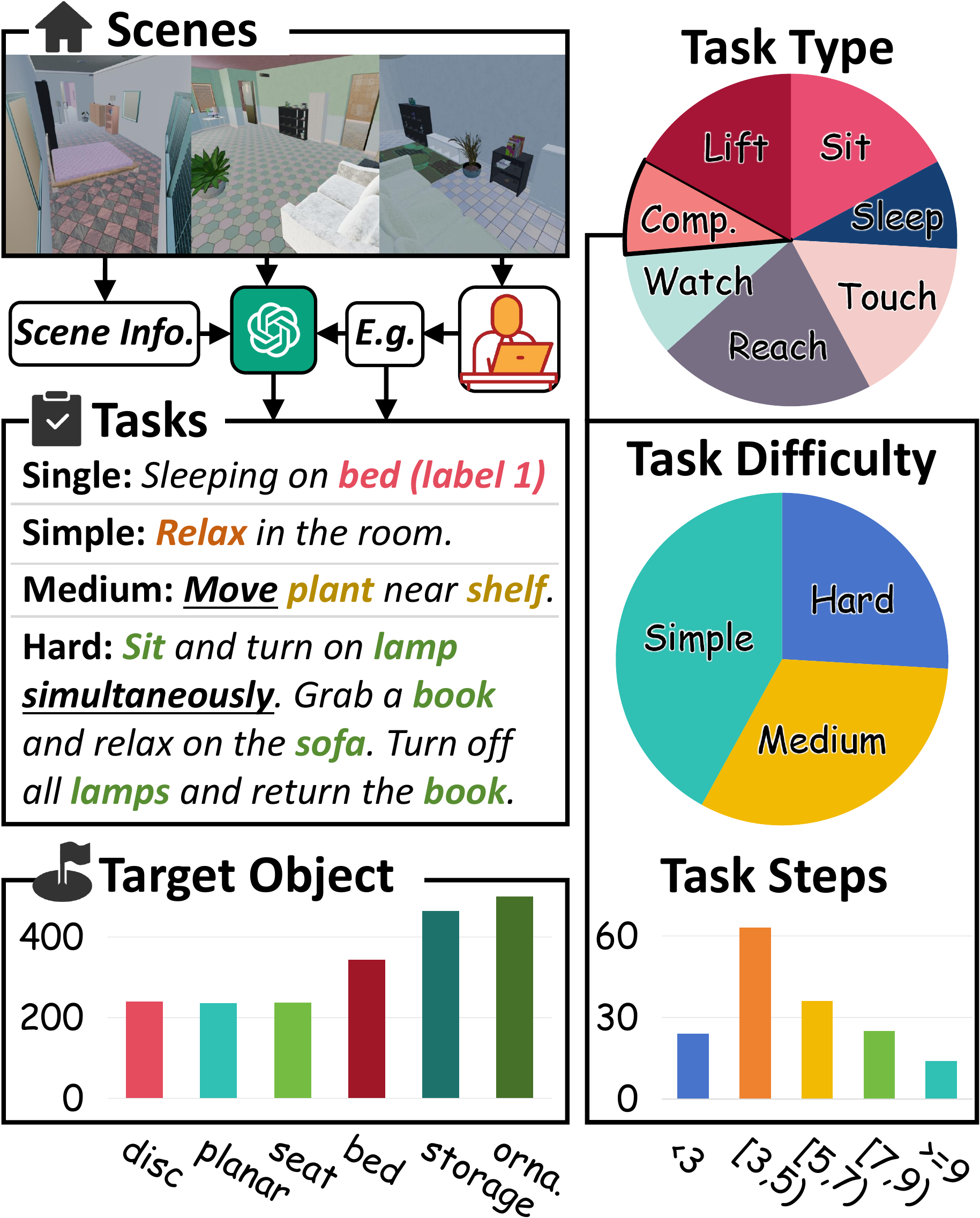}
  \vspace{-6mm}
  \caption{Summary of the random generated scene dataset. The tasks are constructed by a semi-automatic approach. The dataset contains various object categories, task types and difficulties, evaluating a wide range of interaction abilities of humanoid agents.}
  \vspace{-10mm}
  \label{fig:fig4}
\end{wrapfigure}

\textbullet~\textit{Watch} evaluates the understanding of object orientation. It success if the agent stands in front of the target and facing to it, with an angular error of less than $\pi/6$.

 \textbullet~\textit{Sit} \& \textit{Sleep} evaluate interaction with objects of varying shapes and functions. Sit success if the hips are within $0.1$m of the seat area, subject to an upward force, and the torso faces forward. Sleep success if the legs, arms, and torso are all within $0.1$m of the sleep area, with the average force directed upward and the torso facing upward. The angular errors should be less than $\pi/4$.
 
\textbullet~\textit{Touch} is successful if the correct hand is within $0.1$m of the target, and subject force.

\textbullet~\textit{Lift} evaluates dynamic object manipulation. Success if the target being lifted more than $0.25$m above the ground.

We additionally introduce composite tasks, each consisting of multiple interactions. They succeeds if all interactions are accomplished in the specified order. Composite tasks are categorized into simple, medium and hard. \textbf{1) Simple} tasks consist of $<4$ interactions. \textbf{2) Medium} tasks include user intent understanding (e.g., the room is too dark) and dynamic object manipulation (i.e., lift, transport). \textbf{3) Hard} tasks involve long-horizon ($>10$ interactions) and simultaneous interactions with multiple objects (e.g., sit on sofa and put a hand on the side table).

\textbf{Dataset.} To evaluate BiBo’s ability to accomplish different tasks in physical environments, we randomly construct 100 scenes using InfiniGen~\citep{infinigen2023infinite}. These scenes include different rooms and layouts, containing 73 object categories with randomized parameters controlling their shapes and styles. 
For each scene, we construct 6–18 single interaction and 1–3 composite tasks by a semi-automatic approach. Specifically, as in Fig.~\ref{fig:fig4}, volunteers first annotate task examples, after which the LLM generates tasks conditioned on the examples and scene information. This process results in a total of 1,365 single tasks and 162 composite tasks. All tasks are given through natural language instruction, and evaluated three times under different initial conditions.

\textbf{Comparison Method.} 
As comparison methods, UniHSI~\citep{xiao2023unified} focuses on contact interactions, HumanVLA~\citep{xu2024humanvla} follows the VLA paradigm to address transportation tasks, TokenHSI~\citep{pan2025tokenhsi} trains task tokens to manage multiple skills, and CLoSD~\citep{tevet2024closd} employs motion diffusion, enabling diverse scene interactions. As these methods lack an applicable planner, they use a programmatically generated ground-truth plan, with details provided in Sec.~\ref{sec:com_met}. For BiBo, we evaluated both the ground truth plan and online planning.

\textbf{Result.} As shown in Tab.~\ref{tab:tab1}, BiBo achieves an average success rate of $90.2\%$ in single interaction, and $41.0\%$ in composite task. Compared with other methods, BiBo achieves an average improvement of $12.5\%$ and $29.1\%$ on single interaction and composite tasks, respectively. In online planning, BiBo achieves success rates close to those of the ground-truth plan (within $4.38\%$).

\begin{table}
\centering
\caption{Impact of different components in compiler and executor on the task success rates. Act. and Gen. represents the actual executed motion and previous generated motion. The \textbf{bold} and \underline{underline} represent the best and second-best performance, respectively. The results demonstrate the effectiveness of designs in BiBo.}
\vspace{-2mm}
\scriptsize
\resizebox{1.0\textwidth}{!}{
\begin{tabular}{l|cccccc|ccc}
\hline
\multirow{2}{*}{\textbf{Method (\%)}} & \multicolumn{6}{c|}{\textbf{Single Interaction}} & \multicolumn{3}{c}{\textbf{Composite Task}} \\
\cline{2-7} \cline{8-10}
 & Reach $\uparrow$ & Watch $\uparrow$ & Sit $\uparrow$ & Sleep $\uparrow$ & Touch $\uparrow$ & Lift $\uparrow$ & Simple $\uparrow$ & Medium $\uparrow$ & Hard $\uparrow$ \\
\hline
BiBo (ours, w/o Voting) & \textbf{99.18} & 98.87 & 91.13 & 88.67 & \underline{85.82} & \underline{59.75} & 49.51 & \underline{32.05} & 22.22\\
BiBo (ours, w/o Label) &97.00& 	90.21&	48.59&	46.44&	64.89&	58.73&	26.96&	17.31&	7.94 \\
\hline
BiBo (ours, w/o Act.) & 98.09 & 98.87 & 84.18& 73.78 & 81.80 & 28.34 & 28.43 & 16.02 & 10.32\\
BiBo (ours, w/o Gen.) & 98.64&\underline{99.25}&95.62&\underline{93.78}&84.40&56.58&\underline{57.35}&30.77&\underline{23.81}\\
BiBo (ours, w/o IK)&\underline{98.82}&\underline{99.25}&\textbf{95.96}&92.89&48.94&6.80&31.37&13.46&3.17\\
\hline
BiBo (ours)         & \textbf{99.18} & \textbf{99.62} & \underline{95.84} & \textbf{94.89} & \textbf{86.05} & \textbf{65.42} & \textbf{58.82} & \textbf{36.54} & \textbf{27.78} \\
\hline
\end{tabular}
}
\label{tab:tab2}
\vspace{-2mm}
\end{table}

\begin{table}
\centering

\caption{Comparison between \textbf{text-to-motion} methods on the HumanML3D. We evaluates efficiency, motion quality, and physical plausibility. $\uparrow$, $\downarrow$ and $\rightarrow$ means the higher, smaller and closer to the ground truth is preferred. Arbi. and Phys. indicate the support for arbitrary-length and physical plausible generation. The \textbf{bold} and \underline{underline} represents the best and second-best performance.} 
\vspace{-2mm}
\scriptsize
\setlength{\tabcolsep}{1.5pt}
\resizebox{\textwidth}{!}{
\begin{tabular}{l | c | c c c c c c |  c c c c}
\hline
\multirow{2}{*}{\textbf{Method}} & \textbf{Efficiency} & \multicolumn{5}{c}{\textbf{Motion Quality}} & & \multicolumn{4}{c}{\textbf{Physical Plausibility}} \\
\cline{2-2} \cline{3-8} \cline{9-12}
 & AITS $\downarrow$ & Arbi. & FID $\downarrow$  & R.P.@1 $\uparrow$ & R.P.@2 $\uparrow$ & R.P.@3 $\uparrow$ & Diversity $\rightarrow$ & Phys. & Pen. $\downarrow$ & Float $\downarrow$ & Skate $\downarrow$ \\
\hline

\textit{Ground Truth}                                & -     &-& 0.001       & 0.514 & 0.706 & 0.800& 9.503&  \ding{55} & 0.00 & 22.9 & 0.21\\
\hline
MDM  \citep{tevet2022human}                          & 24.74 &\ding{55}& 0.423      & 0.406 & 0.603 & 0.719& \underline{9.559}&  \ding{55} & \underline{0.15} & 28.6 & 0.33 \\ 
MotionLCM  \citep{dai2024motionlcm}                  & \textbf{0.042} &\ding{55}& \underline{0.072}      & 0.510 & 0.703 & 0.797& 9.598& \ding{55} & 0.65 & 27.4 & 0.81 \\
MotionStreamer \citep{xiao2025motionstreamer}       & 2.584 &\ding{51}& 0.084     & 0.432 & 0.615 & 0.716& \textbf{9.549} &\ding{55} & 0.17 & 23.2 & 0.68 \\
MoGenTS  \citep{yuan2024mogents}                     & 0.836 &\ding{55}& \textbf{0.046}      & \underline{0.521} & \underline{0.713} & \underline{0.804}& 9.617&\ding{55} & 0.80&26.2&0.96 \\
MoConVQ   \citep{yao2024moconvq}                   & -     &\ding{55}& 3.279      & 0.309 & 0.504 & 0.614& 8.010 &\ding{51} & 0.25 & 32.0 & \underline{0.29} \\
DiP \citep{tevet2024closd}                     & 0.257 &\ding{51}& 0.210      & 0.466 & 0.653 & 0.754& 9.570 &\ding{55} & \textbf{0.14} & 24.5 & 0.65\\
CLoSD      \citep{tevet2024closd}               & 2.873  &\ding{51}& 2.861       & 0.367 & 0.553 & 0.665& 8.256 &\ding{51} & 0.30 & \textbf{20.2} & \textbf{0.01}\\
\hline

BiBo (ours)  & \underline{0.047} &\ding{51}& 0.076      & \textbf{0.542} & \textbf{0.738} & \textbf{0.829}& 9.606 &\ding{55} &  0.32 & 25.3 & 0.74\\
BiBo (ours, Phy.) & - &\ding{51}& 1.883 & 0.411 & 0.604 & 0.716& 8.298 &\ding{51} & 0.19 & \underline{20.6} & \textbf{0.01}\\
\hline
\end{tabular}
}
\label{tab:text2motion}
\vspace{-5mm}
\end{table}

\begin{wraptable}{t}{0.45\textwidth}
\centering
\caption{Comparison of \textbf{control accuracy} across methods using MAE. $\downarrow$ means smaller is better. \textbf{Bold} is the best performance.}
\vspace{-2mm}
\scriptsize
\setlength{\tabcolsep}{3pt}
\begin{tabular}{l c c c}
  \toprule
\textbf{Method}  & Head $\downarrow$ & Hand $\downarrow$ & Foot $\downarrow$  \\
\midrule
DiP~\citep{tevet2024closd}  & 0.0663 & 0.0830 & 0.0540\\
MotionLCM~\citep{dai2024motionlcm}  & 0.0952 & 0.1470 & 0.0955\\
\hline
BiBo (ours)   & \textbf{0.0310} & \textbf{0.0571} & \textbf{0.0335}\\
BiBo (ours, w/o LDM)  & 0.0705 & 0.0918 & 0.0608\\
\bottomrule
\end{tabular}
\label{tab:con}
\vspace{-1mm}
\end{wraptable}

Experimental results demonstrate that effectiveness of designs in BiBo, revealing the potential of general-purpose VLMs in controlling humanoids.

\textbf{Ablation Study.} We conduct ablation studies to validate the proposed designs in BiBo. Specifically, in the compiler, we introduce a voting mechanism, and leverage image labels to facilitate the visual reasoning. In the executor, we generate future motion condition on the actual executed motion and previous generated motion, and apply IK for precise control.
The results in Tab.\ref{tab:tab2} show that Voting and Label improve task success rate by $4.1\%$ and $22.9\%$.
IK improves the accuracy of joint control, while executed and generated motions affect physical adaptability and precise interaction, respectively.

\subsection{Motion Quality}

\textbf{Setting.} We evaluate the motion quality from both quantitative and qualitative perspectives. On the quantitative side, we report Average Inference Time per Sequence (AITS) reflects the computational overhead~\citep{chen2023executing}; Fréchet Inception Distance (FID), R Precision and Diversity evaluate the motion fidelity and diversity~\citep{guo2022generating}; Penetration, Float, and Skate reflect the physical plausibility~\citep{yuan2023physdiff}; Mean Absolute Error (MAE) evaluates the control precision.
On the qualitative side, we conduct human evaluations and visual inspections, and further perform case studies that analyze both successful and failure cases.

\textbf{Dataset.} We adopt the HumanML3D dataset. Following common practice, we use motion sequences whose lengths fall inside $[40, 200]$ frames. A total of 24,545 motion episodes paired with 66,633 motion captions from the training split are used to train the motion diffusion executor, while 4,646 motion episodes paired with 12,536 captions from the test split are reserved for evaluation.

\textbf{Comparison Method.} We adopt state-of-the-art methods for both fixed-length and arbitrary-length generation, including physical and non-physical approaches. Details are provided in the Sec.~\ref{sec:com_met}.

\textbf{Quantitative Result.} As in Tab.~\ref{tab:text2motion}, BiBo handles on-the-fly control ($>20$Hz), and demonstrates advantages (non-physical $+3.5\%$ and physical $+7.3\%$ relatively) in text alignment (R.P.) across comparison methods. It improves motion realism (FID) in real-time arbitrary-length generation by $63.8\%$ relatively, demonstrates comparable physical plausibility to CLoSD. As in Tab.~\ref{tab:con}, BiBo achieves the highest control precision.
These results demonstrate that the proposed executor provides an effective medium for bridging general-purpose VLM with the physical world.

\textbf{Ablation Study.} We employ LDM and causal attention to mitigate the discontinuity between future and current motion, and use previous generated motion and the actual executed motion to promote smoothness and environmental awareness. 
We conduct ablation studies to demonstrate the contribution of these designs. Specifically, discontinuity can be manifested as an abrupt change in joint velocities. As a result, we quantify discontinuity by computing average joint acceleration $\bar{a} = \E_j(\|\vp_j^{n+1} + \vp_j^{n-1} - 2\vp_j^n\|_2)~/~t^2$ at the initial frame $n$ of the future motion, where $\vp_j^n\in\sR^3$ is the position of joint $j$ at frame $n$, and $t$ is the frame interval in seconds.

\begin{table}[t]
  \centering
  \begin{minipage}[t]{0.38\textwidth}
    \centering
    \caption{\textbf{Motion discontinuity} evaluated by average joint acceleration $\bar{a}$.}
    \vspace{-1mm}
    \scriptsize
    \setlength{\tabcolsep}{10pt}
    \begin{tabular}{l | c}
      \hline
      \textbf{Method ($m^2/s^2$)} & \textbf{$\bar{a}$} \\
      \hline
      CLoSD~\citep{tevet2024closd} & 0.0610 \\
      \hline
      BiBo (ours, w/o LDM)     & 0.0879 \\
      BiBo (ours, w/o Causal)  & 0.0626 \\
      BiBo (ours, w/o Gen.)     & 0.0698 \\
      BiBo (ours, w/o Act.)     & \textbf{0.0370} \\
      BiBo (ours)              & 0.0379 \\
      \hline
    \end{tabular}
    \label{tab:jerk}
    \vspace{3mm}
\setlength{\tabcolsep}{5pt}
    \caption{Number of preferred motions or interactions in \textbf{User study}.}
    \vspace{-1mm}
    \scriptsize
\begin{tabular}{l | c}
  \hline
\textbf{Method} & \textbf{Count}   \\
\hline
DiP(CLoSD)~\citep{tevet2024closd} & 20 \\
MotionLCM~\citep{dai2024motionlcm} & 53 \\
\hline
BiBo (ours) & \textbf{77} \\
\hline
\end{tabular}
\label{tab:usr}
  \end{minipage}
  \hfill
  \begin{minipage}[t]{0.58\textwidth}
    \centering
    \caption{Impact of different components in BiBo on motion quality, evaluated using HumanML3D. $\uparrow$ and $\downarrow$ indicate higher and smaller is preferred, respectively. Phys. shows support for physical plausibility. \textbf{Bold} indicates the best performance.}
    \vspace{-1mm}
    \scriptsize
    \setlength{\tabcolsep}{5pt}
    \renewcommand{\arraystretch}{1.2}
    \begin{tabular}{l | c | c | c c c }
      \hline
      \textbf{Method} & \textbf{Phys.} & \textbf{FID} $\downarrow$  & \multicolumn{3}{c}{\textbf{R.P.@$1\sim3$}} $\uparrow$ \\
      \hline
      \textit{Ground Truth} & \ding{55} & 0.001 & 0.514 & 0.706 & 0.800 \\
      \hline
      BiBo (ours, w/o LDM) & \ding{55} &0.238 & 0.467 & 0.662 & 0.762 \\
      BiBo (ours, w/o LDM, Phy.) & \ding{51} &2.138 & 0.376 & 0.561 & 0.674 \\
      BiBo (ours, w/o Causal) & \ding{55} & 0.101 & 0.526 & 0.716 & 0.801 \\
      BiBo (ours, w/o Causal, Phy.) & \ding{51} & 2.377 & 0.376 & 0.571 & 0.679 \\
      BiBo (ours, w/o Gen., Phy.) & \ding{51} &2.312 & 0.382 & 0.577 & 0.689 \\
      BiBo (ours, w/o Act., Phy.) & \ding{51} & 1.414 & 0.419 & 0.616 & 0.721 \\
      \hline
      BiBo (ours) & \ding{55} & \textbf{0.076} & \textbf{0.542} & \textbf{0.738} & \textbf{0.829} \\
      BiBo (ours, Phy.) & \ding{51} & 1.883 & 0.411 & 0.604 & 0.716 \\
      \hline
    \end{tabular}
    \label{tab:text2motion_ablation}
  \end{minipage}
\end{table}

According to the results in Tab.~\ref{tab:con},~\ref{tab:jerk} and ~\ref{tab:text2motion_ablation}, both LDM and causal attention improve motion quality. Incorporating previous generated motion effectively reduces discontinuity during motion transitions. Incorporating actual executed motion may affect generation quality, but enhance adaptability to environmental interactions as in Tab.~\ref{tab:tab2}.

\begin{figure}[t!]
    \centering
    \vspace{-1mm}
    \includegraphics[width=1.0\textwidth]{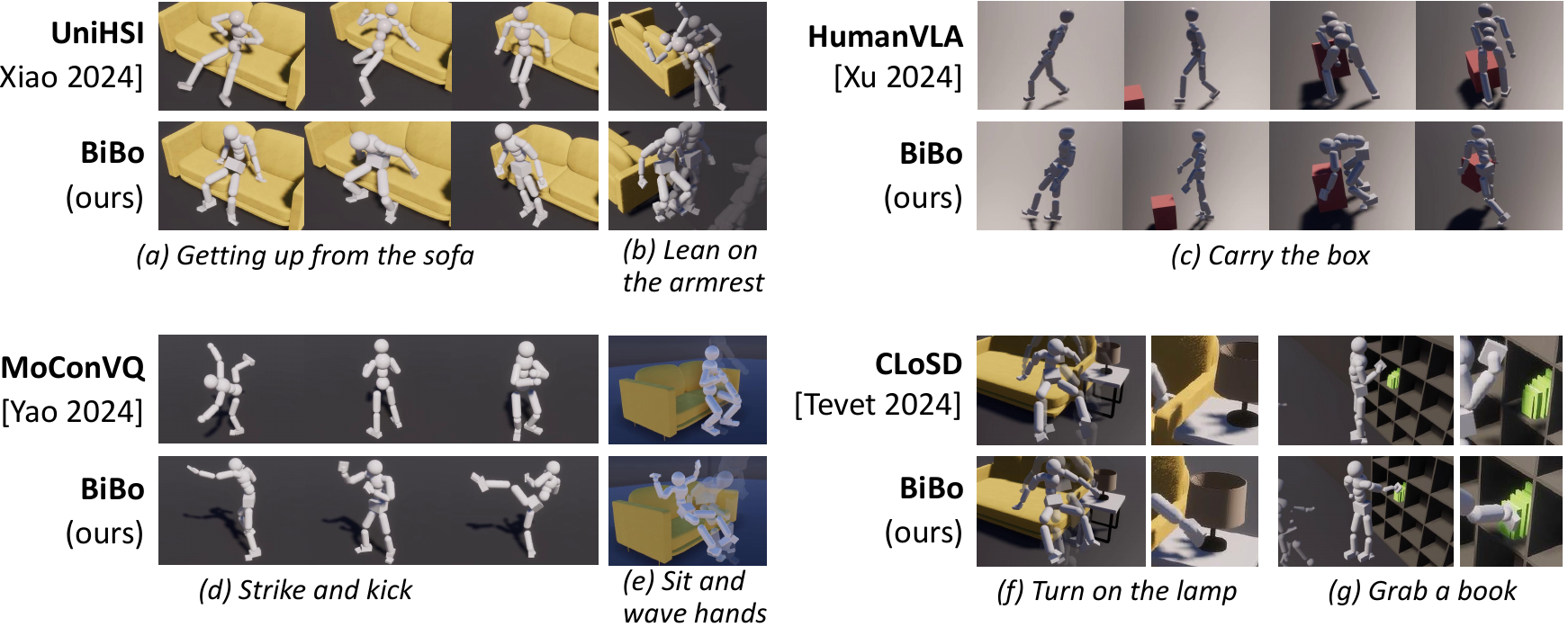}
    \vspace{-5mm}
    \caption{Visualization of executing results of comparison methods. Compared with BiBo, UniHSI generates less natural motions, while HumanVLA requires stricter initial positioning for transportation. MoConVQ shows limited motion activity, and CLoSD struggles to achieve precise control.}
    \label{fig:fig5} 
    \vspace{-6mm}
\end{figure}
\textbf{Qualitative Result and Case Study.} We conduct a user study to evaluate generation quality through questionnaires. The questionnaire consists of 5 pairs of motions and scene interactions generated by BiBo, MotionLCM, DiP(CLoSD) under the same prompts. Each pair of motions is displayed in the same row with the left–right order randomized, and participants are asked to select the one with higher generation quality. The selections from 30 volunteers are summarized in Tab. \ref{tab:usr}. The motions and scene interactions generated by BiBo are preferred.

We conduct visual evaluation with the comparison methods in Fig.\ref{fig:fig5}, including MoConVQ, HumanVLA, UniHSI, and CLoSD.
In (a) and (b), UniHSI produces unnatural movements, whereas BiBo generates natural standing and leaning motion. In (c), when initial agent pose is not aligned with the target, HumanVLA fails to pick up the box, while BiBo performs human-like locomotion and completes the task. In (d) and (e), BiBo accurately follows the text prompt to perform the strike and raise hand actions. For (f) and (g), BiBo achieves higher control precision compared with CLoSD.

We further visualize how three strategies for extending future motion respond to environmental feedback: (1) relying only on previously generated but not executed motion, (2) relying only on previously executed motion, and (3) integrating both by using LDM. In the experiment, we place a desk in front of the agent, and instruct it to raise a hand and then slap downward, which lead to a hand–desk collision. 

\begin{figure}[t!]
    \centering
    \includegraphics[width=1.0\textwidth]{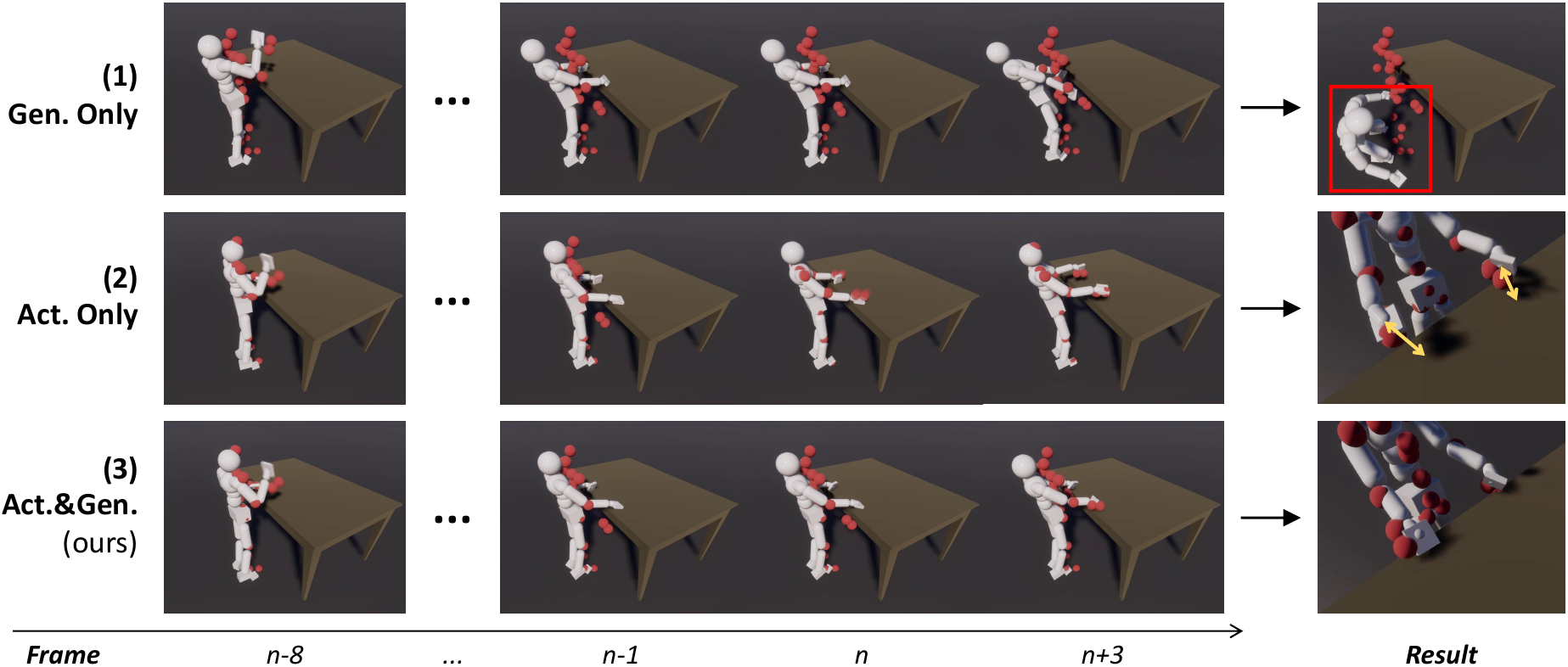}
    \caption{Visual comparison between the executed result of different motion generation method, where the red balls in the image represents the generated motion. Act. and Gen. denotes extending future motion from actual executed motion and previous generated motion, respectively. Extending only from the generated motion fails to account for physical feedback, which may lead to falls. In contrast, extending only from the executed motion introduces discontinuities, resulting in jitter. Our method addresses both issues by incorporating physical feedback while avoiding discontinuities.}
    \label{fig:fig6}     
    \vspace{-1.5em}
\end{figure}

As in Fig.~\ref{fig:fig6}, for (1), the generated future motion (represented by the red balls) ignores the desk collision, continuing to drive the hand downward and ultimately causing the agent to lose balance. For (2), the generated future trajectory (frame $\geq n$) exhibits discontinuous jumps relative to the preceding motion (frame $< n$), producing jitter that bounces the hand off the desk surface. By comparison, our method in (3) preserves motion continuity between frames $n-1$ and $n$, while adapting to physical collision by gradually redirecting the hand upon the desk surface at frame $n+3$. This smooth transition reduces jitter and allows the hands to maintain contact with the table.

%% file: sections/5_conclusion.tex
\section{Conclusion}

We introduce BiBo, a framework that empowers off-the-shelf Vision-Language Models to control humanoid agents. Our key insight is that VLMs can control humanoid agents without costly data collection or task-specific training. To achieve this, BiBo comprises two novel components, an embodied instruction compiler that compiles high-level instructions into executable commands, and a diffusion motion executor that generates motions consistent  with the physical environment. Experiments show that BiBo achieves high success rates across multiple task designs and maintains high text-to-motion fidelity while performing complex interactions.

\textbf{Limitations and Future Work.} First, our executor is trained on a text-to-motion dataset of limited size, which may restrict its generalization capability. With the availability of larger-scale motion datasets~\citep{motionx,motionmillion}, there is potential to further enhance robustness and generalization. Second, while our model incorporates environmental feedback through motion execution results, explicitly modeling environmental geometry—such as height maps~\citep{cen2024generating} or basis point set~\citep{yi2024generating} features—remains an important direction for future exploration. Third, we focus on human–scene interactions in this paper, but there is potential to extend our framework to broader interaction modes, such as hand–object interactions~\citep{chao2018learning} and human–human interactions~\citep{liang2024intergen}. We leave these directions to future studies.

%% file: sections/A_sup_related_work.tex
\section{Supplemantary Related Work}
\subsection{Motion Diffusion Model}
Diffusion~\citep{ho2020denoising,song2020denoising} has emerged as a powerful generative framework for motion synthesis~\citep{khani2025motion}. It improves generation diversity compared to variational autoencoders (VAEs)~\citep{petrovich2021action} and generative adversarial networks (GANs)~\citep{shiobara2021human}, while being more efficient than GPT-like next-token prediction~\citep{zhang2023generating} and BERT-like masked modeling~\citep{guo2024momask} methods. While pioneering works directly denoised Gaussian noise into motion sequences~\citep{tevet2022human,liang2024intergen,yi2024generating}, recent approaches extend diffusion with gradient guidance~\citep{xie2023omnicontrol} and ControlNet~\citep{gang2025strong} for controllability, latent diffusion models (LDMs)~\citep{xiao2025motionstreamer,hong2025salad} for enhanced motion quality, and flow matching (Consistency Models)~\citep{dai2024motionlcm,jiang2025motionpcm} for faster inference. In this work, we adopt LDM architecture with Classifier Free Guidance and few-step denoising, supporting real-time control while achieving high-fidelity.

\subsection{Large Vision-Language-Action Model} 

Large Vision Language Models (VLMs) have demonstrated cross-domain generalization capability~\citep{achiam2023gpt,team2023gemini,bai2023qwen}, enabling them to perform action planning for embodied agents conditioned on environmental context and user instructions~\citep{yao2024moconvq,xiao2023unified,wang2024sims}. Among these methods, Vision-Language-Action~\citep{ma2024survey,zhong2025survey} attracts extensive research attention. They typically learn action token by finetuning VLM on collected action data~\citep{kim2024openvla,black2024pi_0,black2025pi_}, thereby driving low-level action executor. Unlike low degree of freedom (DoF) platform (e.g. robot arms, vehicles)~\citep{brohan2022rt,brohan2023rt,zhou2025opendrivevla}, humanoids process higher dimensional action space, requiring a strong executor and extensive finetuning data~\citep{bjorck2025gr00t,ding2025humanoid}. BiBo explores an alternative approach to bypass finetuning for action tokens. It employs an embodied instruction compiler that guides an off-the-shelf VLM to output structured action commands. These commands drive an executor trained on open-source human motion dataset, thereby performing diverse scene interaction.

%% file: sections/B_datasets.tex
\section{Datasets}
\label{sec:data}
\subsection{Random Generated Scene Dataset}
\begin{figure}[t!]
    \centering
    \includegraphics[width=1.0\textwidth]{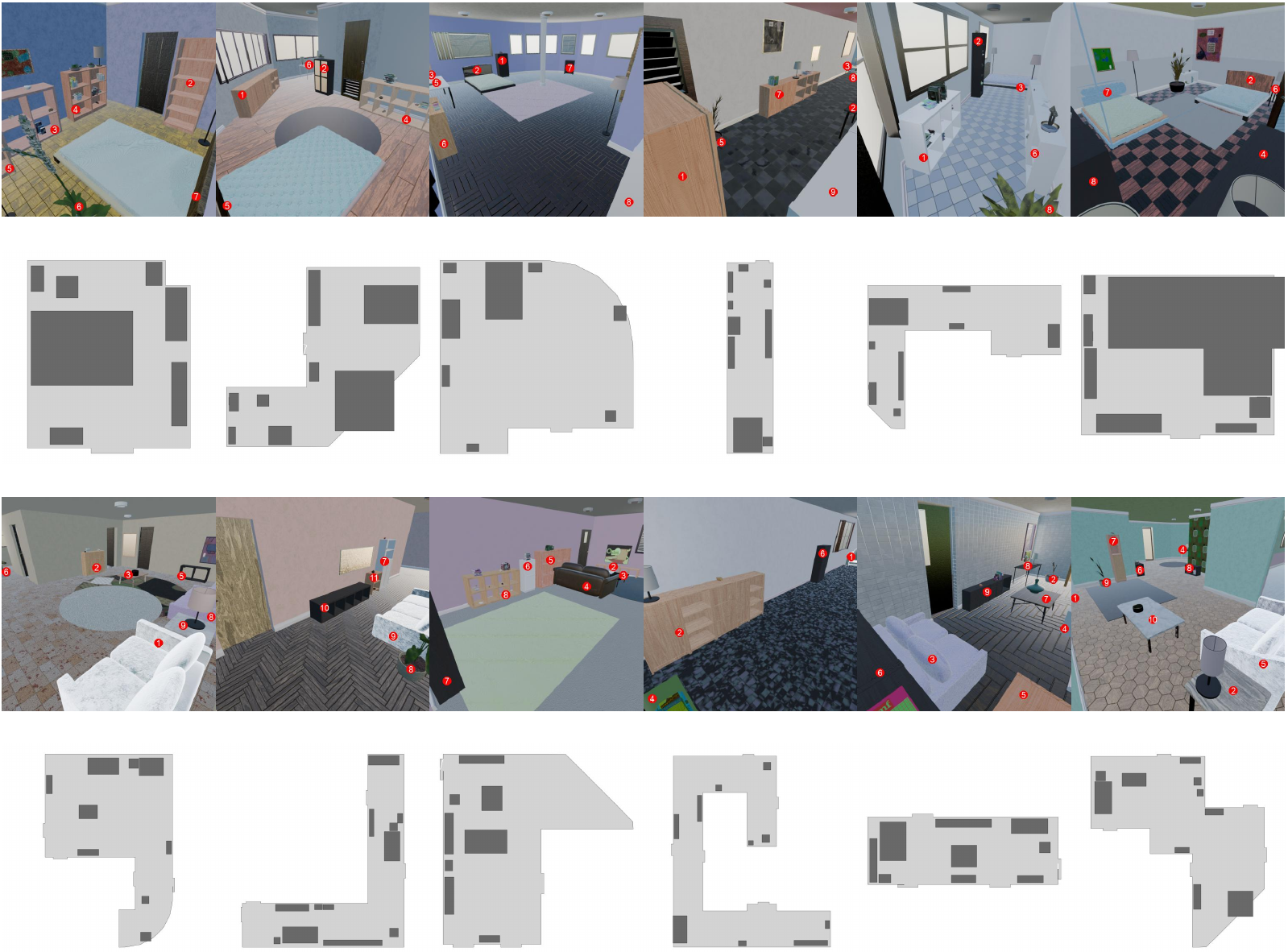}
    \caption{Visualization of the randomly generated scenes and corresponding floor plans.
The top two rows show bedroom scenes, and the bottom two rows show living room scenes. In the scene images, each object is annotated with a red label. In the floor plans, light gray regions denote the floor, while dark gray regions represent the objects.}
    \label{fig:fig7} 
    \vspace{-6mm}
\end{figure}
\subsubsection{Statistics}
The scene dataset comprises 100 scenes spanning diverse room types and layouts, including 50 living rooms and 50 bedrooms. The floor areas range from $12.30$ to $99.18\mathrm{m}^2$, with an average of $49.12\mathrm{m}^2$, featuring various floor plan geometries as illustrated in Fig.~\ref{fig:fig7}. The objects include diverse categories such as ornaments (e.g., plant containers, trinkets), containers (e.g., shelves, cabinets), tables (e.g., TV stands, desks), planar surfaces (e.g., monitors, wall panels), as well as furniture (e.g., sofas, beds) and miscellaneous items. The distribution of these categories is shown in Fig.~\ref{fig:fig4}.

Each scene contains 6–18 single-interaction tasks and 1–3 composite tasks, resulting in a total of 1,365 single tasks and 162 composite tasks. The single-interaction tasks include 297 sit, 150 sleep, 282 touch, 367 reach, 177 watch, and 294 lift tasks. Notably, reach tasks are also embedded within other interaction tasks to improve testing efficiency. The composite tasks comprise 68 simple, 52 medium, and 42 hard cases, with the length distributions shown in Fig.~\ref{fig:fig4}. Examples of tasks are shown in Tab.~\ref{fig:fig7}. The evaluation criteria for task success are described in Sec.~\ref{sec:task_completion}.

\subsubsection{Scene Construction}
\label{sec:data:scene:scene}
The scenes are generated using InfiniGen\footnote{https://infinigen.org/}~\citep{infinigen2023infinite}, which produces Blender-format \texttt{\small{\textbf{.blend}}} scene files. For each generated scene, all light sources and cameras are removed to ensure a consistent simulation environment. Each object in the scene, including walls and furniture, is then exported as an \texttt{\small\textbf{.obj}} mesh file representing its visual geometry, and decomposed into convex collision bodies using the VHACD algorithm~\citep{mamou2016volumetric}. The visual meshes and their corresponding collision bodies are subsequently organized into a unified \texttt{\small\textbf{.urdf}} asset file. These URDF files, along with metadata such as object positions and orientations, are imported into IsaacGym Preview 4\footnote{https://developer.nvidia.com/isaac-gym} to construct the physical simulation environment. The initial position of the humanoid agent is dynamically determined at runtime to introduce spatial diversity and prevent initialization bias.

The scene and object generation process follows InfiniGen’s default indoor generator, with minor modifications to facilitate task construction. Specifically, the footrests are removed from sofas and blankets are removed from beds. Object extraction is conducted through the Blender Python API 4.2.0\footnote{https://docs.blender.org/api/current/index.html}, and all meshes are simplified to fewer than 10,000 faces to reduce computational overhead. The simplified meshes are subsequently processed using Trimesh\footnote{https://trimesh.org/} for vertex filtering and mesh repair, ensuring watertight topology and the removal of abnormal geometries.

During the import process, since the built-in VHACD and SDF collision modules in IsaacGym exhibit limitations in geometric approximation and compatibility, we employ PyVHACD\footnote{https://github.com/thomwolf/pyVHACD}
 to manually construct the collision bodies. Given an object mesh as input, VHACD decomposes it into up to $64$ convex hulls that approximate the object’s external geometry. Each convex hull is further abstracted by its axis-aligned bounding box (AABB), which is encoded into the URDF as a geometry box element to define the collision shape. Decorative items such as trinkets and plants are defined as dynamic assets, with a density of approximately 50kg/m³, comparable to wooden boxes. Large furniture such as shelves and walls are defined as static assets to stabilize the simulation and reduce computational cost.

To determine the humanoid agent’s initial position during evaluation, Shapely\footnote{https://github.com/shapely/shapely} is used to construct 2D polygons representing the floor and scene objects. After applying a uniform padding operation to prevent boundary collisions, a random point located inside the floor polygon but outside all object polygons is sampled as the agent’s starting position. This ensures valid initialization across diverse layouts while preventing overlap with scene geometry.

\subsubsection{Task Generation}
\label{sec:data:scene:task}
\textbf{Task Format.} All tasks are stored in JSON \texttt{\small\textbf{.json}} format, as shown in Lst.~\ref{lst:json-schema}. It contains two fields: \texttt{\small\textbf{prompt}} and \texttt{\small\textbf{mission}}. The \texttt{\small\textbf{prompt}} field is a natural language instruction of the task, and the  \texttt{\small\textbf{mission}} field is a two-level array of interaction objectives.
The first level represents interactions that must be completed sequentially, 
and the second level represents interactions that can be completed simultaneously. 
Each interaction objective includes the target object name \texttt{\small\textbf{target}} and the interaction type \texttt{\small\textbf{type}}. There are five available interaction types, including {\small\itshape[watch, sit, sleep, touch, lift]}.
Note that the {\small\itshape reach} task is implicitly included in these interaction types, since the agent should first navigate to the target object before performing the corresponding interaction.

The object name follows the format \texttt{\small\textbf{category[index][*]}}. 
The \texttt{\small\textbf{category}} component is required and specifies the category of the target object, 
wildcard matching all unvisited objects of that type in the scene. 
The \texttt{\small\textbf{[index]}} component is optional and designates a specific object instance. 
The \texttt{\small\textbf{[*]}} component is also optional and indicates that, after this interaction, 
the object will not be excluded from subsequent wildcard matching. 

\begin{figure}[t]
\centering
\begin{minipage}{\linewidth}
\begin{lstlisting}[language=json, caption={Example of a task JSON schema. The first object name does not contain an index, indicating that it can correspond to any \textit{bookstack} or \textit{bookcolumn} in the scene. It contains an (\textbf{*}), indicating that the target object of the last interaction can be the same as that of the first interaction. The two intermediate interaction objectives are placed in the same second-level array, representing that they must be achieved simultaneously.}, label={lst:json-schema}]
{
  "prompt": "Grab a book, then sit on the couch while turning on the light, and finally place the book somewhere.",
  "mission": [
    [{"object":"book*","type":"touch"}],
    [{"object":"sofa1","type":"sit"}, {"object":"lamp1","type":"touch"}],
    [{"object":"book","type":"touch"}]
  ]
}
\end{lstlisting}
\end{minipage}
\end{figure}

\textbf{Single Interaction.} We use a rule-based program to construct single-interaction tasks.
For each interaction type, the program queries the scene for eligible interactive objects and generates corresponding prompts accordingly.
Specifically, the target objects for the watch task are directional planar surfaces such as TV screens or wall paintings, with example prompts like {\small\itshape``watching TV1.''}
The sit task targets seating furniture such as sofas or beds, e.g., {\small\itshape``sitting on sofa1.''}
The sleep task targets beds.
The touch task involves ornamental objects in the room, with example prompts such as {\small\itshape``grab a book from bookstack1''} or {\small\itshape``turn on lamp1.''}
The lift task targets large ornaments or containers, with prompts like {\small\itshape``lifting large plant container1.''}

\textbf{Composite Task.} The composite tasks are constructed using a semi-automatic pipeline, where volunteers manually annotate 5 simple, 3 medium, and 3 hard task examples. The remaining tasks are then generated by GPT-4o based on the provided examples and scene information. All generated tasks are manually reviewed to ensure that they are achievable. The task difficulty is categorized according to the criteria described in Sec.~\ref{sec:task_completion}.

Composite tasks are composed of multiple single interactions that occur either sequentially or concurrently. Compared with single interactions, they feature more diverse interaction patterns, such as {\small\itshape``laying on the coffee table''}, {\small\itshape``leaning on the bookshelf''}, or {\small\itshape``sitting while turning on the lamp''}.

\begin{mytbox}[label={box:composite_prompt}]{Prompt for Composite Task Generation}
\textbf{SYSTEM\textcolor{red}{*}:} You are an intelligent task generator.
Your goal is to create interaction tasks that a humanoid agent can perform within a given scene.
\\
\\
Each task should be in JSON format with two fields: 

- prompt: natural language instruction of the task, can be both abstract and concrete

- mission: a two-level array of interaction objectives. The first level is sequential, while the second level is simultaneous.
\\
\\
Each interaction objective contains two fields:

- type: one of [watch, sit, sleep, touch, lift]

- target: name of target object "category[index][*]". "category" is required. [index] specifies a particular object (optional), omit to match all uninteracted objects of that category. [*] indicates this object should not be omitted in subsequent matching.
\\
\\
Task difficulty criteria:

- simple: contain $<4$ steps (including navigating to another object between two interactions, e.g., "sit on sofa1 and then sit on sofa2" = 3 steps)

- medium: 4-10 steps, or contains dynamic object manipulation (e.g., lift, transport)

- hard: $>10$ steps, or contains simultaneous interactions with multiple objects (e.g., sit on sofa and put a hand on the side table)
\\
\\
\textbf{USER\textcolor{red}{*}:} Example of simple task:

[examples]

Example of medium task:

[examples]

Example of hard task:

[examples]
\\
\\
The multi-view scene images are provided, each image contain object labels, corresponding to:

- 1: sofa (-0.4, 5.1)

- 2: desk (-0.2, 3.8), containing: [nature shelf trinket1]

- 3: monitor (1.1, 5.3)
\\
\\
You are currently generating an simple task, with abstract prompt and 2 interactions. No simultaneous interaction. No dynamic object manipulation. Please analyze before outputting the final task, and enclose your final answer in $>>>$ and $<<<$.
\end{mytbox}

During task generation, the GPT-4o is provided with several components: example tasks, Blender-rendered multi-view scene images with object labels, and a list of objects with their corresponding coordinates and parent–child relationships. To ensure a balanced difficulty distribution, we use a random number generator instead of GPT-4o to determine task difficulty. Specifically, we explicitly specify the task difficulty to be generated and the corresponding criteria to be satisfied. 

To capture multi-view scene images, we first compute the positions of all wall corners based on the floor polygon. Cameras are then placed along the bisectors of these corners at a height of 2 m, capturing images with a 30° downward tilt and a 75° field of view (FOV). For placing object labels, the point cloud of each object is first projected onto the camera plane to generate a mask. Then we use OpenCV\footnote{https://opencv.org/} to compute the distance from each pixel within the mask to its boundary, and the pixel with the maximum distance is selected as the center of the object label.

Based on these inputs, it outputs tasks in JSON format. The generated JSON file is then parsed and automatically validated by querying the scene to check (1) whether the target objects exist, (2) whether the interaction type belongs to the predefined interaction list, and (3) whether the coordinates of simultaneously contacted target objects are within a 1m distance. Finally, all generated tasks are manually reviewed to ensure correctness and executability. The prompt is shown in Box.~\ref{box:composite_prompt}.

\subsection{HumanML3D}
\subsubsection{Statistics}
We train the diffusion-based motion executor on the HumanML3D\footnote{https://github.com/EricGuo5513/HumanML3D} dataset~\citep{guo2022generating}, which comprises 14,616 human motion sequences paired with 44,970 natural language descriptions. HumanML3D covers everyday activities, spatial interactions, and complex body dynamics. The motion data are extracted from HumanAct12~\citep{guo2020action2motion} and AMASS~\citep{mahmood2019amass}, augmented through mirroring to double the dataset size, and normalized into a 263-dimensional relative-coordinate motion representation. 

The dataset is divided into 24,843 motion sequences for training and validation and 4,382 for testing. After selecting motion sequences with lengths ranging from 40 to 200 frames, a total of 24,545 motion episodes paired with 66,633 captions are used for training and validation, while 4,646 episodes paired with 12,536 captions are reserved for testing.

\subsubsection{Motion Representation}
We follow \citet{tevet2022human}, each motion episode is represented as a sequence of joint positions, rotations and velocities in 3D space, sampled at 20 frame per second (FPS). Each motion frame $x \in \mathbb{R}^F$ encodes a single pose and is defined as:
\[
x = (\dot{r}_a, \dot{r}_x, \dot{r}_z, r_y, \mathbf{j}_p, \mathbf{j}_r, \mathbf{j}_v, \mathbf{f}),
\]
where $\dot{r}_a$ is the root angular velocity around the Z-axis; $\dot{r}_x$ and $\dot{r}_z$ are root linear velocities in the XY-plane; $r_y$ is the root height. The joint features include local joint positions $\mathbf{j}_p \in \mathbb{R}^{3(J-1)}$, rotations $\mathbf{j}_r \in \mathbb{R}^{6(J-1)}$, and velocities $\mathbf{j}_v \in \mathbb{R}^{3J}$, all defined relative to the root. Additionally, $\mathbf{f} \in \mathbb{R}^4$ denotes binary foot contact indicators for four foot joints (two per leg).

%% file: sections/C_compiler.tex
\section{Embodied Instruction Compiler}
\label{sec:emb_ins_com}
\subsection{Online Control Loop}
The embodied instruction compiler takes as input the user’s instruction and the scene observation, and outputs the next structured action command for the executor to perform. To achieve this, Sec.~\ref{sec:met:com} introduces a three-stage VQA pipeline designed to progressively fill in the structured action command. In implementation, the embodied instruction compiler can be further divided into three modules:
(1) an action Planning Module, which includes the off-the-shelf VLM and the three-stage VQA process;
(2) an humanoid agent state machine controlled by the Planning Module; and
(3) a Navigation Module that provides path-planning services for the state machine.
These three modules, together with the executor and the physical environment, form an online control loop for the humanoid agent.

Specifically, the Planning Module is invoked when a new user instruction arrives, when the previous interaction is completed, or when 150 frames have elapsed since the module was last invoked. It reads the current user instruction, scene information, and agent state (including all the already executed action commands and the ongoing action), and decides whether to skip, start a new action, or end the current one. When a new action starts, the system applies the three-stage VQA process to generate the action command, which is then passed to the state machine.

The state machine is consist of navigation and interaction states. Upon receiving an action command, it switches between navigation and interaction according to the interaction type, the target object, scene information, and the agent’s current state (position and ongoing action). In the navigation state, it outputs action commands based on the Navigation Module’s path-planning results; in the interaction state, it issues action commands that drive the executor to synthesize the next interaction.

During navigation, the Navigation Module first constructs a pixel obstacle map of the scene and then performs path planning using a modified A* algorithm. In our implementation, we introduce an additional repulsion term in the A* cost function to guide the trajectory away from obstacles and ensure safe navigation.

\subsection{VLM-based Action Planning Module}
\subsubsection{Off-the-shelf Vision Language Model}

BiBo employs GPT-4o, accessed via the official API\footnote{https://platform.openai.com/}
.
In the Python environment, we utilize the OpenAI API library\footnote{https://github.com/openai/openai-python}
 and manually construct both the conversation-prompt framework and the chain-of-thought procedure.
BiBo can also be integrated with other variants of VLMs (e.g., Qwen~\citep{bai2023qwen,bai2025qwen2}, Claude~\citep{anthropic2024claude3}, and Gemini~\citep{team2023gemini}).
In our experiments, we compare the scaling capabilities of VLMs of different sizes and examine how tailored prompt designs influence models of comparable capacity.
Specifically, we employ Claude 3.5 Sonnet\footnote{https://www.claude.com/platform/api}
, Qwen2.5-VL\footnote{https://qwen.ai/apiplatform}
, and GPT-4o mini as large-, medium-, and small-scale comparison methods, respectively.

\begin{mytbox}[label={box:inst_refine}]{User Instruction Refinement}
\textbf{SYSTEM\textcolor{red}{*}:} You are a state-of-the-art intelligent humanoid robot. 
Given a vague command, you can interpret it into a sequence of clear, executable instructions according to the scene.
\\
\\
Definition of Clear Instruction:

- Imperative sentence. 

- Have exactly one verb. 

- If the verb denotes an interactive action, it must explicitly state the object. If not stated in the origin command, you should choose one from the object list.

- The object stated in the sentence must exists in the provided object list.

- Make sure every verb in the original command is included in the instruction sequence.

- Don’t miss details like adjectives, directions, and numbers.
\\
\\
\textbf{USER\textcolor{red}{*}:} Vague Command: Jump in place. Sit, and sleep.

Objects:

- 1: sofa1 (5.0 ,5.0)

- 2: simple bookcase1 (6.0, 4.5), containing [book column1, book column2]

Clear Instructions:
\\
\\
\textbf{ASSISTANT\textcolor{red}{*}:} Jump. Sit on the sofa1. Sleep on bed1.
\\
\\
\textbf{USER:} \underline{\textless Multi-view Images\textgreater}

Vague Command: \underline{\textless User Instruction\textgreater}.

Objects: 
\underline{\textless Scene Description\textgreater}

Clear Instructions:
\end{mytbox}

\begin{figure}[t!]
   \centering
    \includegraphics[width=1.0\textwidth]{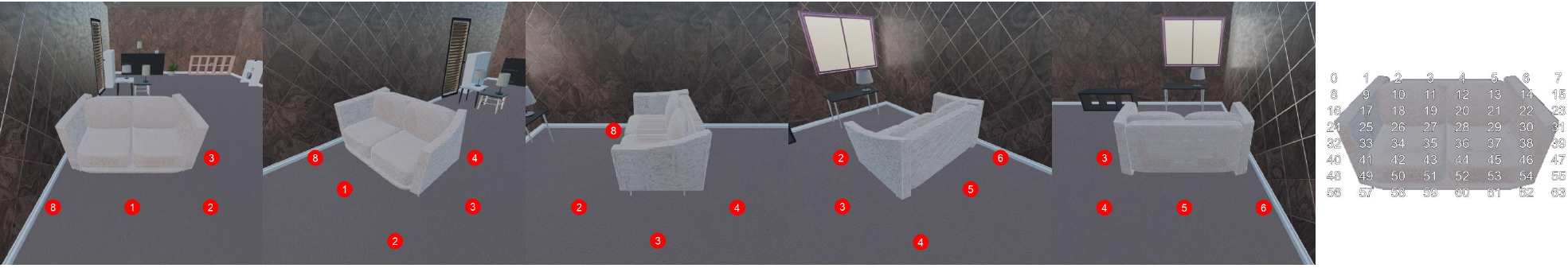}
    \caption{Visualization of the labeled images: the left side presents multi-view images with labels indicating directions relative to the object, while the right side shows the labeled image used for determining target position of key joints.}
    \label{fig:fig8} 
\end{figure}

\begin{figure}[t!]
  \centering
  \begin{minipage}[t]{0.57\textwidth}
    \centering
    \includegraphics[width=\linewidth]{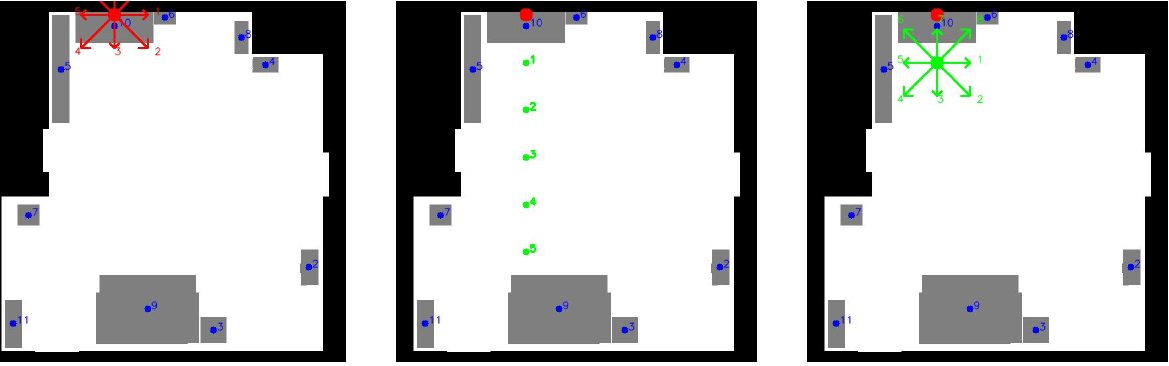}
    \caption{Visualization of agent pose reasoning process and the BEV image inputs for VLM.}
    \label{fig:fig9}
  \end{minipage}
  \hfill
  \begin{minipage}[t]{0.38\textwidth}
    \centering
    \includegraphics[width=\linewidth]{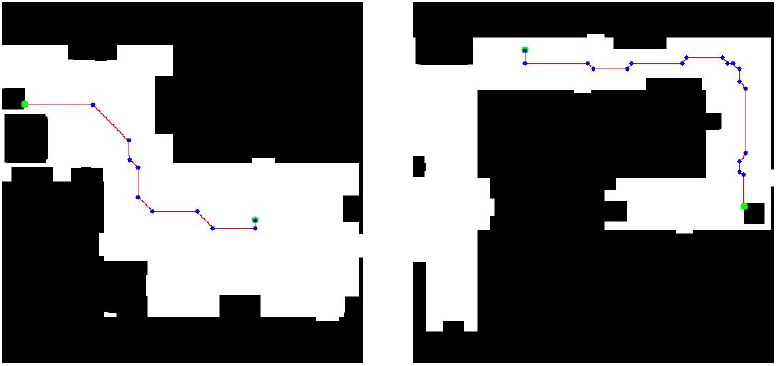}
    \caption{Visualization of the navigation result of the modified A*.}
    \label{fig:fig10}
  \end{minipage}
\end{figure}

\subsubsection{Details of Basic Attribute Analysis}
\label{sec:com:pln:base}
In Basic Attribute Analysis, the compiler takes as input the user instruction, scene information, and the agent’s status. The VLM refines the user instruction based on these inputs as in Box.~\ref{box:inst_refine}, and decides whether to \texttt{\small\textbf{skip()}}, \texttt{\small\textbf{start()}} a new action, or \texttt{\small\textbf{end()}} an ongoing action, using the prompt in Box.~\ref{box:next_plan}. When an action starts, the VLM first generates a brief summary of the action as the motion caption. Then, it analyzes the basic motion attributes according to the scene objects and the agent’s current state, facilitating the subsequence reasoning process, using the prompt in Box.~\ref{box:basic_analysis1} and ~\ref{box:basic_analysis2}.

\begin{mytbox}[label={box:next_plan}]{Planning for The Next Action}

\textbf{SYSTEM\textcolor{red}{*}:} You are a state-of-the-art intelligent humanoid robot.

You are currently in a scene with the following objects, and the layout of the scene is provided in the image:

\underline{\textless Multi-view Images\textgreater}

\underline{\textless Scene Description\textgreater}

You should perform a sequence of actions to fulfill the following instructions, step by step:

\underline{\textless User Instruction\textgreater}
\\
\\
In each step, the sensor will provide you with your position, the objects you are holding, your current states and action elapsed time. You should decide your next action command accordingly, including:

1. skip(): Stay in the current state.

2. start(action): Start doing the specified action. Will add the action to your current state, which means you are doing the action.

3. stop(action): Stop doing the specified action. Will remove the action from your current state, which means you are no longer doing the action.
\\
\\
Example of actions:

1. Jumping

2. Leaning against wall1

3. Taking hammer1 from toolbox1

4. Standing on the right of table1 three meters away
\\
\\
Constraints:

- Each step you can only give one command.

- Each action should be indivisible, which containing no more than one verb.

- For interactive actions, you must explicitly specify the target object, as example 2, 3, 4.

- Multiple actions in current state means performing multiple actions at the same time.

- You are encouraged to directly output the command without any additional explanation.
\\
\\
\textbf{USER\textcolor{red}{*}:} Scene: \underline{\textless Scene Description\textgreater}

Position: origin (0, 0) 

State: [(``Standing'', 3s, done)]

Holding: \{left hand: none, right hand: none\}
\\
\\
\textbf{ASSISTANT\textcolor{red}{*}:} stop("Standing.")
\\
\\
\textbf{USER:} \textcolor{gray}{$\dots$(Multi-round Conversation)}

\end{mytbox}

\begin{mytbox}[label={box:basic_analysis1}]{Basic Motion Attribute Analysis (Object)}
\textbf{USER\textcolor{red}{*}:} You are a state-of-the-art intelligent humanoid robot, \underline{\textless VLM Instance ID\textgreater}. The following command describes an interaction with objects marked with ``\textless'' and ``\textgreater''.
Please identify the roles of these objects. 
\\
\\
The roles include ``target'', ``by'', and ``at''. 

- ``target'' is the intented interaction target.

- ``by'' means the interaction is done by using these objects.

- ``at'' means the interaction is done at these objects.
\\
\\
Please answer in the following format:

~~~~target: \textless target\textgreater

~~~~by: [\textless object1\textgreater, \textless object2\textgreater, ...]

~~~~at: [\textless object3\textgreater, \textless object4\textgreater, ...] 
\\
\\
Example:

Question 1: 
Grab the \textless cloth1\textgreater from the \textless washing machine1\textgreater.

Answer 1: 

~~~~target: \textless cloth1\textgreater

~~~~at: \textless washing machine1\textgreater

Question 2:
Turn on the \textless light1\textgreater with the \textless switch1\textgreater.

Answer 2:

~~~~target: \textless light1\textgreater

~~~~by: \textless switch1\textgreater
\\
\\
Sentence: \underline{\textless Motion Caption\textgreater}
\end{mytbox}

\begin{mytbox}[label={box:basic_analysis2}]{Basic Motion Attribute Analysis (Motion)}
\textbf{SYSTEM\textcolor{red}{*}:} You are a state-of-the-art intelligent humanoid robot, \underline{\textless VLM Instance ID\textgreater}. You can assess the details of an action to perform it accurately. 
\\
\\
The details include:

- Interaction Type: one of [contact, non-contact, long-range, manipulation]. Contact refers to contact with the target object (if present), non-contact denotes proximity without contact, long-range indicates interact at a distance, and manipulation involves moving the object.

- Key Joints: the most important joints involved in the interaction, available options: ['pelvis', 'left\_foot', 'right\_foot', 'left\_hand', 'right\_hand', 'head']

- Use Inverse Kinematic: true or false, enable for precise or stable interactions (e.g., touching small objects, carrying), disable for high dynamic motions (e.g., dancing)
\\
\\
Constraints:

- No more than two contact points. If there may be more, choose the most important two.

- One hand cannot manipulate multiple objects.

- You are encouraged to directly output the details without any explanation.
\\
\\
\textbf{USER\textcolor{red}{*}:} Interaction: Turn on \textless lamp1\textgreater on \textless table1\textgreater.

State: You are holding book1 in your right hand.

Details:
\\
\\
\textbf{ASSISTANT\textcolor{red}{*}:}- Interaction Type: contact

- Key Joints: [``left\_hand'']

- Use Inverse Kinematic: true
\\
\\
\textbf{USER:} Interaction: \underline{\textless Motion Caption\textgreater}

State: You are \underline{\textless Agent State\textgreater}.

Details:
\end{mytbox}

\textbf{Scene Information.} It includes multi-view images and a corresponding scene description. The multi-view images are captured in the same manner as described in Sec.~\ref{sec:data:scene:task}, by capturing images from all wall corners to maximize the coverage of the scene contents. Each image is labeled with object IDs. The scene description contains object IDs, coordinates, and parent–child relationships among objects. The parent–child relationships are computed using a union–find algorithm~\citep{cormen2022introduction} based on the containment relationships among the objects’ 2D polygonal shapes, and the object with the largest (or similar) area and the lowest height is selected as the parent.

\textbf{Agent Status.} It includes all previous planning results and executed actions, the currently executing action, the agent’s pose, nearby objects, and objects held by the agent. The previous planning results are provided through multi-round conversation history, while the currently executing action is identified by its motion caption, elapsed time, and execution result. All objects are specified by their names and corresponding IDs.

\textbf{Motion Attributes.} They include the related objects, the key joints involved, the interaction type, and whether IK should be enabled. The related objects are categorized into three types: {\small\itshape target}, {\small\itshape at}, and {\small\itshape by}. The {\small\itshape target} is the target object of interaction, {\small\itshape at} refers to the anchor used for agent localization, and {\small\itshape by} denotes the tool being used. For example, take a book {\small\itshape (target)} from the shelf {\small\itshape (at)} or turn on the TV {\small\itshape (target)} by the remote {\small\itshape (by)}. The {\small\itshape target} is chosen as the anchor if there are no {\small\itshape at} objects. The key joints include the {\small\itshape head}, {\small\itshape hands}, {\small\itshape feet}, and {\small\itshape pelvis} (the latter being controlled by the agent’s position and facing direction). The interaction types are divided into four categories: {\small\itshape contact}, {\small\itshape non-contact}, {\small\itshape distal}, and {\small\itshape manipulate}. {\small\itshape contact} involves physical contact (e.g., sitting), {\small\itshape non-contact} refers to proximity without contact (e.g., hand dryer use), {\small\itshape distal} denotes remote interactions without approaching the target, and {\small\itshape manipulate} refers to interactions that change the object’s state (e.g., carrying).

\subsubsection{Details of Agent Pose Reasoning}
\label{sec:com:pln:pose}
In Agent Pose Reasoning stage, the VLM take the rendered scene images with labels as input, and sequentially infers:
(1) the agent’s direction relative to the anchor object,
(2) the agent’s distance relative to the anchor object, and
(3) the agent’s facing direction.
The success of this reasoning process depends on the choice of scene viewpoint as well as the placement and annotation of the labels within the rendered images.

\begin{mytbox}[label={box:agent_to_anchor_dir}]{Inferring Agent’s Direction Relative to Anchor}
\textbf{USER\textcolor{red}{*}:} You're the state-of-the-art intelligent humanoid robot. When you interact with an object, you can determine your own position relative to it.
\\
\\
The scene layout is provided in the images, which are photos of the \underline{\textless Anchor Object\textgreater} taken from different perspectives. The photos contain red circular markers labeled with integer indices, each marker represents a direction relative to the \underline{\textless Anchor Object\textgreater}. In different photos, the marker with the same index represents the same direction.

\underline{\textless Multi-view Images\textgreater}
\\
\\
Marker Directions: \underline{\textless Direction Label Descriptions\textgreater}

\textcolor{gray}{For example:}

\textcolor{gray}{- Label 1 is in the front of monitor and desk}

\textcolor{gray}{- Label 2 is in the front-left of monitor and desk}

\textcolor{gray}{$\dots$}

When performing \underline{\textless Motion Caption\textgreater}, which direction should you stand in? Your answer should be a marker index enclosed in \textgreater \textgreater \textgreater and \textless  \textless  \textless . 
\end{mytbox}

\textbf{Agent’s Direction Relative to Anchor.} We first capture multi-view rendered images of both the anchor and target objects. The images are rendered with an object-centered setup at a 30° depression angle, sampled at 45° intervals with a 75° field of view, and occluded views are excluded, as shown in Fig.~\ref{fig:fig8}.

Then, we employ Orient Anything\footnote{https://github.com/SpatialVision/Orient-Anything}~\citep{wang2024orient}, a model fine-tuned on DINOv2~\citep{oquab2023dinov2} that can predict object orientation and corresponding confidence scores in a zero-shot manner. For each view, Orient Anything outputs an estimated orientation, which we discretize into directional bins. The confidence scores of all views within each bin are summed, and the bin with the highest aggregated confidence is selected as the object’s front direction. If the overall confidence is below a threshold, the object is considered to have no clear orientation. 

Subsequently, eight labels are uniformly placed around the anchor object at 0.4 m intervals and 45° spacing, excluding non-traversable locations. These labels are projected onto the anchor’s multi-view rendered images and provided to the VLM, together with textual descriptions indicating their spatial relations (if applicable) to the anchor and target, e.g., “Label 1 is in front of the monitor.” The VLM then selects one label as the agent’s relative direction to the anchor object. We use the prompt in Box.\ref{box:agent_to_anchor_dir}. For non-distal interactions, the prompt explicitly states that the label represents a standing location rather than a direction to skip the next reasoning step, as non-distal interaction is expected to occur within a 0.4m range around the object.

\begin{mytbox}[label={box:agent_to_anchor_dst}]{Inferring Agent’s Distance from Anchor}
\textbf{USER\textcolor{red}{*}:} You're the state-of-the-art intelligent humanoid robot. When you interact with an object, you can determine your own position relative to it.
\\
\\
The provided image is a bird-eye-view map of the scene. 

\underline{\textless BEV\textgreater}
\\
\\
Red markers indicate the \underline{\textless Target Object\textgreater}, while blue markers denote the IDs of scene objects. These objects are as follows: 

\underline{\textless Scene Description\textgreater}
\\
\\
The green markers in the image represent a set of candidate standing locations: \underline{\textless Distance Label Descriptions\textgreater}

\textcolor{gray}{For example:}

\textcolor{gray}{There distance to the monitor is:}

\textcolor{gray}{- Label 1 : 0.5m}

\textcolor{gray}{- Label 2 : 1.5m}

\textcolor{gray}{$\dots$}
\\
\\
When you are performing \underline{\textless Motion Caption\textgreater}, which position should you stand in? Your answer should be a marker index enclosed in \textgreater \textgreater \textgreater and \textless  \textless  \textless .
\end{mytbox}

\textbf{Agent’s Distance from Anchor.} As shown in Fig.~\ref{fig:fig9}, we construct a simplified bird eye view (BEV) of the scene using the polygons of floorplan and objects, where walkable areas and obstacles are filled with different colors, and the positions of the agent and objects are indicated by specific labels. Along the previously predicted direction, a series of distance labels are placed starting from 0.5m and spaced at 1m intervals, excluding positions that are not reachable. The prompt template is shown in Box.~\ref{box:agent_to_anchor_dst}, which provides each label’s distance and direction relative to target objects (if target is distinct from the anchor). The VLM selects one label as the final standing location. For non-distal interactions, this step is skipped, and the distance is fixed at 0.4m.

\begin{mytbox}[label={box:agent_facing}]{Inferring Agent’s Facing Direction}
\textbf{USER\textcolor{red}{*}:} You're the state-of-the-art intelligent humanoid robot. When you interact with an object, you can determine your facing direction.
\\
\\
The provided image is a bird-eye-view map of the scene. 

\underline{\textless BEV\textgreater}
\\
\\

Red markers indicate the \underline{\textless Target Object\textgreater}, while blue markers denote the IDs of scene objects. 

\underline{\textless Scene Description\textgreater}
\\
\\

The green arrows in the image represent a set of candidate facing directions:

\underline{\textless Facing Label Descriptions\textgreater}

\textcolor{gray}{For example:}

\textcolor{gray}{- Arrow 1: 0°, facing directly to the monitor.}

\textcolor{gray}{- Arrow 5: 180°, facing away from the monitor.}

\textcolor{gray}{$\dots$}
\\
\\
When you are performing \underline{\textless Motion Caption\textgreater}, which direction should you face? Your answer should be an arrow index enclosed in \textgreater \textgreater \textgreater and \textless  \textless  \textless .
\end{mytbox}

\textbf{Agent’s Facing Direction.} Using the BEV map, we place arrows around the standing location as candidate facing directions, as shown in Fig.~\ref{fig:fig9}. We use the prompt in Box.~\ref{box:agent_facing}. In the prompt, for each candidate we provide (i) its angle (relative to a global reference) and (ii) the object it points toward, restricting objects to those relevant to the current interaction. The VLM then selects one candidate as the final facing direction. The BEV input is omitted at this stage for non-distal interactions. 

\begin{mytbox}[label={box:joint_pos}]{Inferring Target Position of Key Joints}
\textbf{SYSTEM\textcolor{red}{*}:} You are a state-of-the-art intelligent humanoid robot. When interacting with an object, you can determine the target position of a specific joint.
This process involves two steps: (1) locating a target point on the target object, and (2) specifying the joint’s position relative to that target point (including its direction and distance).
\\
\\
Example 1: use the laptop

Joint: right\_hand

Target Point: keyboard

Direction: up

Distance: 0
\\
\\
Example 2: use the hand dryer

Joint: right\_hand

Target Point: outlet

Direction: down

Distance: 0.2 m
\\
\\
\textbf{USER\textcolor{red}{*}:} The image shows a \underline{\textless View Direction\textgreater} view of the \underline{\textless Target Object\textgreater}.
It contains an 8×8 grid of numbered labels (indexed from 1 to 64), with each label corresponding to a point on the surface of the \underline{\textless Target Object\textgreater}.

\underline{\textless Image\textgreater}
\\
\\
You are performing \underline{\textless Motion Caption\textgreater}.
Please identify the target position of your \underline{\textless Joint Name\textgreater} in the following format:

- Target Point: a noun or noun phrase (can include adjectives and other modifiers to better describe its features) describing a part of the \underline{\textless Target Object\textgreater} that your \underline{\textless Joint Name\textgreater} refers to.

- Label: one label index in the image corresponding to the target point.

- Direction: the direction of your \underline{\textless Joint Name\textgreater} relative to the target point. Available options:
  [up, down, left, right, directed into the image, directed out of the image, toward the object center, along the surface normal].
  
- Distance: the distance of your \underline{\textless Joint Name\textgreater} from the target point, using meter (m).
\end{mytbox}

\subsubsection{Details of Key Joint Generation}
\label{sec:com:pln:joint}
This process generates joint target positions relative to the object surfaces, as in Box.~\ref{box:joint_pos}.
Given the agent’s location relative to the object, we select the scene image rendered from the corresponding viewpoint and uniformly place an 8×8 grid of labels on it, as shown in Fig.~\ref{fig:fig8}. Each label corresponding to a point on the object surface.
For each key joint, the VLM selects one label as an anchor point, from which it infers the direction and distance between the joint and the anchor. These values collectively define the joint’s target position.

For contact and manipulate types of interactions, we adopt a simplified strategy.
For objects with a size smaller than 0.25 × 0.25 × 0.25m, the key joint generation process is skipped, and the target position is directly set at the object center.
For larger objects, the reasoning for direction and distance is omitted. Instead, the model determines whether the agent should exert force on the object. If not, the target position is placed on the object surface; if so, it is positioned toward the object center, guiding the agent to apply force to the object.

\subsubsection{Multiple Action Merging}
\label{sec:com:pln:merge}

For actions occurring concurrently (e.g., when one action has not yet ended while another starts), the system first checks for potential conflicts after performing basic attribute analysis.
Conflicts are defined under the following conditions:
\begin{itemize}
    \item The two actions involve the same key joint;
    \item Both are non-distal interactions and their target objects are more than 1m apart;
    \item Their anchor objects are more than 1m apart.
\end{itemize}

If no conflict is detected, the system proceeds to generate a new action command through the subsequent pose reasoning and joint generation processes, and updates the existing action command accordingly.

Specifically, when generating a new action command, the system constructs a merged motion caption combining the ongoing and newly initiated motions (e.g., if {\small\itshape sit} has not ended and {\small\itshape turn on the lamp} starts, the merged caption becomes {\small\itshape sit and turn on the lamp}).
During Agent Pose Reasoning, the rendered image centers on all anchor objects, and the merged motion caption is used in the VQA process of VLM.
During Joint Pose Generation, the system regenerates the key joint target positions for the ongoing actions and overwrites them in the corresponding action commands.

All action commands are stored concurrently in the humanoid state machine’s action command list. The state machine’s final output command uses the merged motion caption, adopts the location and facing of the latest generated action command (since the reasoning already considers the merged caption), and directly combines the key joint positions of all action commands.

\begin{mytbox}[label={box:reflect}]{Example of Explicit Insertion of Reflection Results in Conversation History}
\textbf{ASSISTANT:} start("Sitting on sofa1.")
\\
\\
\textbf{USER:} Command execution failed. 

Reason: You are performing "Sleeping on bed1.", and it is impossible to perform "Sleeping on bed1 and sitting on sofa1.", as bed1 and sofa1 are located too far apart. 
\\
\\
Possible solutions: 

- stop "Sleeping on bed1."

- try another action

\end{mytbox}

\subsubsection{Rule-based Reflection}
\label{sec:com:pln:ref}
To enhance the stability of the three-stage VQA process, we incorporate a reflection mechanism. Reflection is triggered in the following cases: (a) failure to follow the expected QA format; (b) conflicts with ongoing actions; and (c) implausible spatial relationships (e.g., a contact positioned too far from the anchor). When an error is detected, the system responds based on the error type: format violations result in a simple rollback, while logical inconsistencies trigger a rollback followed by the explicit insertion of the reflection result into the conversation history, as in Box.~\ref{box:reflect}.

\subsection{Humanoid Agent State Machine}

The State Machine consists of two states, Navigation and Interaction, and maintains an action command list that stores all currently executing actions. During each frame update, the State Machine first converts the positions and directions in each action commands from local coordinate system, defined relative to its anchor object, into global coordinate system. Then, it employs the the method in Sec.~\ref{sec:com:pln:merge} to convert the command list into a merged action command. Next, the system evaluates the difference between the current agent pose (i.e., location and facing) and the target value. 

When the distance to the target is less than 0.5m and the deviation in facing direction is within 45°, the State Machine transitions into the Interaction state and marks the corresponding action command as executing. Once all executing non-distal actions with target objects are completed, the system switches back to the Navigation state.

\subsubsection{Navigation State} 
\label{sec:com:state:navi}
The State Machine first invokes the Navigation Module to generate a planned path. It identifies the point on this path closest to the agent’s current position as the starting point, and then selects the next waypoint that is both nearest to the start and farther than 0.5m away. 

The direction from the current position to this next point is used as the facing direction. If the angular deviation between the current and target facing directions exceeds 45°, the target location is fixed at the current position, and the motion caption is set to {\small\itshape``A person is slowly turning around in place.''} Otherwise, the target location is set to the next waypoint, and the motion caption is set to {\small\itshape``A person is walking.''}
When the distance to the next path point exceeds 1.2m, the moving speed is set to 1m/s. When the distance falls below 1.2m, the speed decreases linearly until it reaches 0m/s at 0.2m. 

The navigation command is further merged with concurrently executing distal, manipulation, or non-interactive actions (i.e., actions without target objects, which are excluded from the second stage of Basic Attribute Analysis and categorized as the Type-V action), allowing navigation and motion execution to proceed simultaneously.

\subsubsection{Interaction State} First, for non-distal motions, the target location is set to a position 0.3m away from the target object, guiding the agent to walk closer to the target. When the positional error drops below 0.2m, the interaction action command start executed. For contact actions involving force exertion, the joint target position is first assigned to the contact point on the object’s surface. Once the agent is within 0.1m of this point, the target position is set to the object center. 

After 60 frames of execution, the system evaluates whether each interaction is done based on predefined rules: a contact action is done if the relevant joint is applying force, is within 0.25m of the target position, and within 0.1m of the target surface; a non-contact action is done when the joint is within 0.1m of the target position; a distal action is done if its duration exceeds 120 frames; and a manipulation is done when the object’s movement exceeds 0.1m while remaining within 0.1m of the hand. The execution duration and completion status of all actions are continuously fed back to the planner to support subsequent decision-making.

\subsection{Navigation Module}
The Navigation Module performs path planning based on the input agent position, target position, scene floorplan, and all object information. First, it constructs a navigation map using the floorplan and the convex decomposition of all objects, while optimizing both the agent and target positions to ensure they are located outside obstacles. It then performs pathfinding using an modified A* algorithm that incorporates a repulsion term to encourage paths farther from obstacles, followed by a polyline simplification step that converts the dense pixel path into sparse waypoints.

\subsubsection{Mapping Module}

For each scene object, the module applies the current position and rotation to transform the coordinates of the 64 convex AABBs obtained from decomposition (refer to Sec.~\ref{sec:data:scene:scene}), and projects them onto the XY-plane. This calculation is GPU-accelerated with PyTorch\footnote{https://pytorch.org/}. The projected polygons are converted into Shapely geometries, padded by 0.1 to form the 2D polygonal outlines of the objects. The floorplan and object polygons are then processed with PyClipper\footnote{https://github.com/fonttools/pyclipper} for clipping, and discretized into a pixel obstacle map. Pixels covered by the floorplan but not by any object represent navigable areas, while the rest correspond to obstacles. The map is discretized to a resolution of $512 \times 512$, scaled isotropically according to the longest axis of the floorplan.

If the agent position lies within an obstacle, the module searches in four directions (up, down, left, right) to find the nearest accessible area, marking all traversed pixels as navigable. If the target position lies within an obstacle, it is shifted outward according to the direction of the corresponding stand location relative to the object. 

Finally, OpenCV is employed to compute an obstacle distance map, which records the distance from each navigable pixel to the nearest obstacle. Specifically, a morphological dilation operation is iteratively applied to the pixel obstacle map, where navigable pixels are assigned a value of 0 and obstacles 1. The number of dilation iterations before a pixel is removed corresponds to its distance to the nearest obstacle. The process terminates once all pixels become 1.

\subsubsection{Path Planning with Modified A*}
The modified A* algorithm utilizes this obstacle distance map for pathfinding. Specifically, a repulsion term is introduced into the cost function, increasing the step cost as the agent approaches obstacles. This design encourages the generation of paths that stay farther from walls, thereby lowering the overall path cost. The complete algorithm is presented in Alg.~\ref{alg:repulsion_astar}.

We use Shapely to simplify the path from dense pixels into sparse waypoints. We then iterate through each pair of waypoints. If the line connecting them maintains a minimum distance greater than 0.5m from the nearest obstacle (check by padding the line into a rectangle with a width of 0.5m and performing a polygon intersection), the intermediate points are bypassed. This process yields the final simplified path.

\begin{algorithm}
\caption{The Modified A* Path Planning with Repulsion Term}
\label{alg:repulsion_astar}
\KwIn{Start point $\vp_{\text{start}}$, goal point $\vp_{\text{goal}}$, obstacle distance map $\mM$, repulsion ratio $\alpha$, repulsion distance $\beta$, scene-to-pixel scaling ratio $\gamma$}
\KwOut{Path $\mathcal{P} = [\vp_1, \ldots, \vp_T]$}

$\mO \leftarrow (\mM == 0)$;

$\mM \leftarrow \text{max}(\beta - \gamma\mM, 0)~/~\beta$;

$\mM \leftarrow (1 - \alpha) \mM  + \alpha$;

Initialize open set with $\vp_{\text{start}}$ and set $g(\vp_{\text{start}}) = 0$ \tcp*{g is current cost}

\While{open set not empty}{
    $\vp_t \leftarrow \arg\min_{\vp} f(\vp)$ \tcp*{f is estimated cost}
    \lIf{$\vp_t = \vp_{\text{goal}}$}{\textbf{break}}
    \ForEach{neighbor $\vp'$ of $\vp_t$}{
        \lIf{$\mO[\vp']$ is true}{\textbf{continue}}
        $g(\vp') \leftarrow g(\vp_t) + \mM[\vp']$; 
        
        $f(\vp') = g(\vp') + \alpha \cdot h(\vp')$ \tcp*{h is heuristic function}
        
        \lIf{$g(\vp')$ improves}{update open set and parent pointer}
    }
}
Reconstruct path $\mathcal{P}$ from parent pointers; \\
\Return $\mathcal{P}$
\end{algorithm}

%% file: sections/D_executor.tex
\section{Diffusion-based Motion Executor}
The motion executor is a translator to translate high-level vlm instructions into low level human motion. The motion executor is a latent diffusion model, which contains a Variational AutoEncoder (VAE)  encoding human motion into low-dimension latent code and a latent diffusion model (LDM) performing a diffusion process on the latent space.

\subsection{Diffusion Architecture.} We adopt a 9-layer Transformer Decoder architecture with skip connections as the diffusion backbone. Each layer uses 4 self-attention heads, with a model hidden size of 256 and an FFN dimension of 1024. Starting by setting the future motion tokens $\mS^T_f$ as Gaussian noise, where $T$ is the total denosing steps. The diffusion timestep $T$ is represented via a sinusoidal positional embedding passed through a small MLP. The diffusion denoiser $\mathcal{F}$ iteratively denoises $\mS^t_f$ using DDIM scheduler, condition on command $\mathcal{C}$ and the latent of preceding actual executed motion $\mS_a$. Specifically, we encode the motion captions using a pretrained text encoder ~\citep{devlin2019bert}, where the [CLS] token is taken as the caption token $\vs_m \in \sR^H$. Other control parameters are encoded with an MLP, and their representations are summed to form the control token $\vs_c$. $\vs_m$ and $\vs_c$ are concatenated with $\mS_a$, together conditioning the denoising process $\mS^{t-1}_f = \mathcal{F}(\mS^t_f, [\mS_a,\vs_m,\vs_c])$
by cross attention with $\mS^t_f$. 

\subsection{Diffusion Process}
Our denoising network follows the DDPM framework with a fixed forward diffusion and a learned reverse denoising process. Let the future-motion latent be denoted by $\mS_f \equiv \mS_f^0$, and let $\{\beta_t\}_{t=1}^T$ be a variance schedule with $\alpha_t \!=\! 1-\beta_t$ and $\bar{\alpha}_t \!=\! \prod_{i=1}^t \alpha_i$.

\paragraph{Forward process}
Starting from clean data $\mS_f^0$, the forward (noising) Markov chain adds Gaussian noise over $T$ steps:
\[
q\!\left(\mS_f^{t}\mid \mS_f^{t-1}\right)
= \mathcal{N}\!\left(\mS_f^{t};\, \sqrt{1-\beta_t}\,\mS_f^{t-1},\, \beta_t\,\mathbf{I}\right),
\]
This can be written in closed form as:
\[
q\!\left(\mS_f^{t}\mid \mS_f^{0}\right)
= \mathcal{N}\!\left(\mS_f^{t};\, \sqrt{\bar{\alpha}_t}\,\mS_f^{0},\, (1-\bar{\alpha}_t)\,\mathbf{I}\right).
\]

\paragraph{Reverse process with CFG}
At sampling time, we sample the terminal noise from a standard normal distribution and iterate backward:
\[
\mS_f^{T} \sim \mathcal{N}(0, 1), \qquad
\mS_f^{t-1} = \frac{1}{\sqrt{\alpha_{t-1}}}
\left(
\mS_f^{t}
-\frac{\epsilon\!\left(\mS_f^{t},\, \vs_m,\, \vs_c\right)}{\sqrt{1-\alpha_{t-1}}}
\right),
\qquad
\mS_f = \mS_f^{0}.
\]
We employ Classifier-Free Guidance (CFG) to modulate the control strength, using the condition set $(\vs_m,\vs_c)$ within the noise predictor $\epsilon(\cdot)$.

\paragraph{Training objective}
The reverse model $\epsilon_\theta$ is trained to predict the injected noise at each timestep via the simplified DDPM objective:
\[
L_{\text{simple}}
=
\mathbb{E}_{t,\, \mS_f^{0},\, \epsilon}
\left[
\left\|
\epsilon - \epsilon_\theta\!\left(\mS_f^{t},\, t,\, \vs_m,\, \vs_c\right)
\right\|^{2}
\right],
\]
where $\mS_f^{t}$ is obtained from the forward process as above. After $T$ denoising steps, the procedure yields the clean future-motion latent $\mS_f=\mS_f^{0}$.
\subsection{VAE Architecture}

The proposed Variational Autoencoder (VAE) utilizes a shared 9-layer Skip-Transformer architecture for both encoder and decoder. Each layer employs 4-head multi-head attention with a model dimension of 256 and a 1024-dimensional feed-forward network. We use the GELU activation function and set the dropout rate to 0.1. To enforce temporal causality, we implement two key mechanisms. First, a causal mask is applied within the self-attention layers to the concatenated sequence of latent and frame tokens. Second, for cross-attention, we introduce an growing memory window that expands its size in correspondence with the target timestep, restricting access to only the relevant prefix of latent variables. These designs effectively prevent future information leakage, ensuring faithful and time-consistent sequence generation.

\subsection{Training Details}
Our implementation uses PyTorch
as the primary deep learning framework for training and inference. We leverage the transformers\footnote{https://huggingface.co/docs/transformers} library for tokenizer and model utilities. The diffusion process is implemented with the diffusers\footnote{https://huggingface.co/docs/diffusers}
toolkit, including noise schedulers and sampling pipelines. All experiments run on a CUDA-enabled backend \footnote{https://developer.nvidia.com/cuda-toolkit}. All training is conducted on the HumanML3D dataset~\citep{Guo_2022_CVPR}. Motion sequences are processed at 20 Hz. Each sequence is concatenated with a stance sequence at the beginning to serve as the initial state, which enable smoother step-to-step transitions during auto regressive learning. When trianing with control conditions, we apply random conditioning by sampling from 60 predefined combinations of control signals including trajectory, heading, velocity, and key joint positions, enabling flexible spatial control during generation. During generation,following CLosdc~\citep{tevet2024closd}, we set prefix length to 20 and generation length 40. We use the AdamW optimizer with parameters $\beta_1=0.9$, $\beta_2=0.999$, weight decay=0, and $\epsilon=10^{-8}$. 
The learning rate is initialized to $5 \times 10^{-4}$ and follows a cosine decay schedule with a linear warm-up over the first 2000 steps. We utilize a DDIM sampler, apply gradient clipping with a threshold of 1.0, and set the unconditional probability to 0.1. For inference, we use 5 DDIM steps and a guidance scale of 7.5. The model is trained for up to 3000 epochs with a batch size of 256. All experiments are run with a fixed random seed of 1234 for reproducibility.
\subsection{IK Post Optimization}
\label{sec:exe:ik}
To ensure plausible hand-object interaction, we introduce an Inverse Kinematics (IK) post-processing step to refine the upper-limb posture when a grasp target is provided by the planner. This step operates on the generated pose for each frame. Our method employs a custom variant of the FABRIK algorithm. Specifically, we solve for a 3-DOF kinematic chain consisting of the left shoulder, elbow, and wrist, with the shoulder joint acting as a fixed root. The segment lengths (upper arm and forearm) are derived from the initial pose and remain constant. Unlike standard FABRIK, our primary IK target is the wrist joint, not the hand end-effector. This design choice provides a stable base for subsequent hand orientation and grasping. The iterative process begins by placing the wrist joint directly at the target position. Then, standard backward and forward passes are executed to satisfy the kinematic constraints of the arm segments. The hand’s position is subsequently reconstructed, where it is placed at a fixed distance from the solved wrist position and oriented along the vector of the forearm. For targets beyond the arm's reachable workspace, the chain is fully extended and pointed towards the target. The entire iterative refinement process is capped at a maximum of 50 iterations, with a tolerance of $10^{-3}$.
\subsection{Tracking Policy}
We utilize the PHC tracking policy~~\citep{luo2021dynamics}, enabling humanoid agents to accurately track the generated joints in physical space. We initialize the policy network with weights from CLoSD~\citep{tevet2024closd}, which is trained using Isaac Gym simulator ~\citep{makoviychuk2021isaac}\footnote{https://developer.nvidia.com/isaac-gym}.

%% file: sections/E_com_methods.tex
\section{Comparison Methods}
\label{sec:com_met}
\subsection{Task Completion}
\subsubsection{UniHSI}
UniHSI~\citep{xiao2023unified} represents each human–scene interaction task as a sequence of contact interactions. It employs AMP~\citep{peng2021amp} achieve contact goals.
Each contact contains a standing position, a contact flag (if not in contact, the agent navigates to the standing position), and a set of contact joint information. Each joint information contains a joint name, a contact normal and 200 surface points on the contact area (which, in the original task definition, correspond to the target object and a specific part index of that object).
During task execution, the agent also needs to obtain nearby scene height information. 

We use the official GitHub repository\footnote{https://xizaoqu.github.io/unihsi/} and the released model checkpoints. During evaluation, we import scenes in IsaacGym using the same asset options as in the scene configuration, and additionally sample each object’s point cloud using Trimesh to construct the contact surfaces and scene height map.
To build the ground-truth task plan, we first procedurally generate contact surfaces and standing points, then create a navigation path based on the scene map, and finally convert it into the UniHSI task format.

\begin{figure}[t!]
    \centering
    \includegraphics[width=1.0\textwidth]{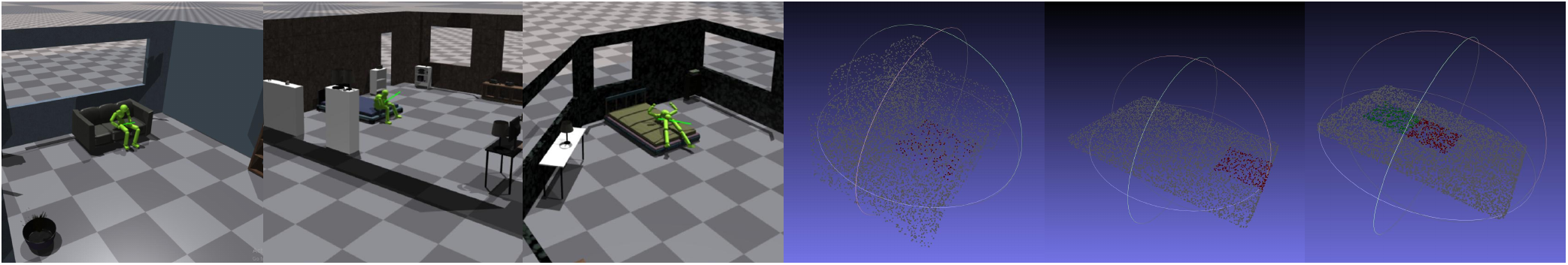}
    \caption{Visualization of task execution process in UniHSI and  the ground truth contact surfaces in random generated scene.}
    \label{fig:fig11} 
\end{figure}

As shown in Fig.~\ref{fig:fig11}, contact surfaces are generated according to object categories and interaction types.
For sofas, we decomposed the backrest, armrests, and seat cushion by point-cloud height, take the 0.5 × 0.5m region at the center front of the cushion as the pelvis-joint contact surface for the Sit task, and set the standing point 0.5m in front of the cushion’s front edge center.
For beds, we identify the mattress, pillow, and headboard based on the height of the point cloud, with contact normals facing upward. For the Sleep task, the central 0.5 × 0.5m region of the mattress is used as the pelvis contact surface, and the pillow’s upper surface point cloud as the head contact surface.
For the Sit task, we use the 0.5 × 0.5m region at the center front of the mattress as the pelvis contact surface, with upward normals and the standing point 0.5m in front of the mattress’s front edge center.
For Touch tasks, where the target objects are small, we use the object’s point cloud as the contact surface for the left or right hand. We uniformly sample 36 candidate standing points along a 0.5m-radius circle around the object and select the first reachable one as the standing position, with the contact normal pointing from the object to the standing point. Tasks of type watch, lift, and composite cannot be executed by UniHSI. Navigation is implemented using the same A* algorithm as in BiBo.

As in Fig.~\ref{fig:fig5}, the qualitative evaluation of UniHSI covers two interaction scenarios: (a) getting up from the sofa and (b) leaning on the armrest. In (a), we use the same configuration as the sitting task, placing the standing point 0.5 m in front of the cushion’s front-edge center at a height of 0.86 m. In (b), UniHSI and BiBo use the same configuration, with the pelvis and hands set to make contact with the upper part of sofa’s armrest.

\begin{figure}[t!]
  \centering
  \begin{minipage}[t]{0.38\textwidth}
    \centering
    \includegraphics[width=\linewidth]{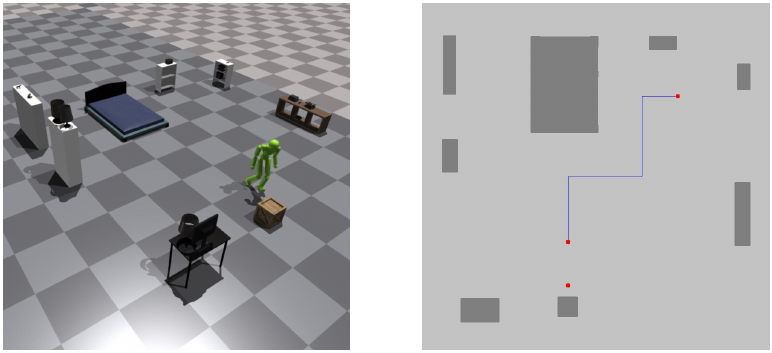}
    \caption{Visualization of HumanVLA and the path planning result.}
    \label{fig:fig12}
  \end{minipage}
  \hfill
  \begin{minipage}[t]{0.57\textwidth}
    \centering
    \includegraphics[width=\linewidth]{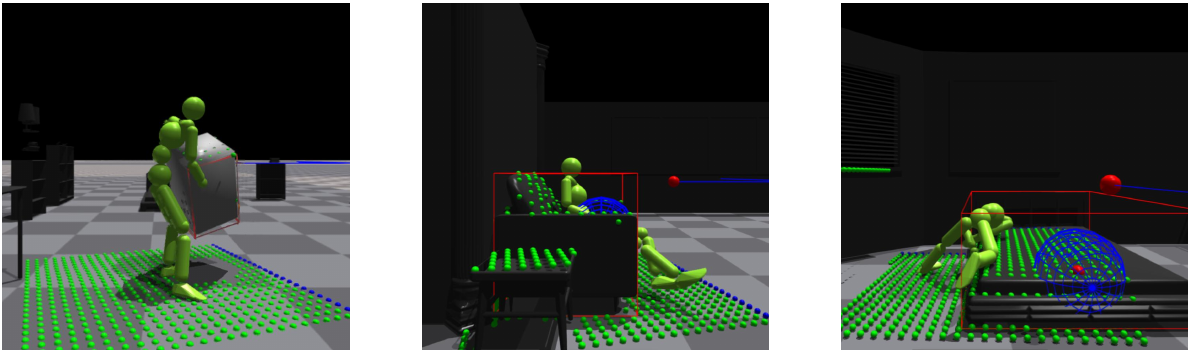}
    \caption{Visualization of the task execution process in TokenHSI, illustrating the height map and target position.}
    \label{fig:fig13}
  \end{minipage}
\end{figure}

\subsubsection{HumanVLA}
HumanVLA~\citep{xu2024humanvla} performs object transportation tasks based on visual observation and natural language descriptions.
Its task format consists of a prompt embedding, the initial and target poses of scene objects, the target object name, and the waypoints before and after transportation. At runtime, the model takes as input the current and target poses of the target object, the next waypoint, the rendered scene image, and the BERT~\citep{devlin2019bert} embedding of the prompt.

We use the official repository\footnote{https://github.com/AllenXuuu/HumanVLA} and the released {\small\itshape HumanVLA-Teacher} model checkpoint (the Student model struggles to complete the task). The randomly generated scenes and tasks are organized according to the HumanVLA task format. Walls are removed, and the asset import options and settings remain consistent with the original scene configuration. Since HumanVLA is sensitive to interaction poses, we modify the A* pathfinding by setting the destination 1.5 m in front of the target object, and adding an extra waypoint 0.5 m in front of the target to encourage a direct approach toward it. The task execution process and pathfinding result are shown in Fig.~\ref{fig:fig12}.

During the qualitative evaluation shown in Fig.~\ref{fig:fig5}, the agent is initialized at a position of (-3m, 2m) relative to the target, facing along the +y-axis. The waypoint is set at (-1m, 0m) relative to the target. The box has size of 0.36 × 0.36 × 0.5m, with 0.5m representing its height.

\subsubsection{TokenHSI}
TokenHSI~\citep{pan2025tokenhsi} achieves multi-interaction generalization and integration by fine-tuning task-specific tokenizers and action heads. Its task format consists of scene objects and a task plan, where each step in the plan is defined by an action type and its corresponding parameters. Specifically, the {\small\itshape traj} action includes a list of path waypoints, the {\small\itshape sit} action specifies the interaction point, seat bounding box, position, and orientation, and the {\small\itshape carry} action defines the target box, its bounding box, and the destination position. During execution, the model also samples a local height map around the agent as part of its observation to enhance environmental awareness.

We adopt the official repository\footnote{https://liangpan99.github.io/TokenHSI/} and the {\small\itshape Longterm 4 Basic Skills} checkpoint, constructing all tasks according to the TokenHSI format. The scene is first converted into an file structure compatible with TokenHSI, which includes each object’s height map, mesh, bounding box, facing direction, up vector, and sit position. When importing scenes, the same asset options as those in the original scene configuration are applied. The constructed tasks follow the same format as in TokenHSI, encompassing object poses and the complete action plan. Trajectory waypoints for {\small\itshape traj} actions are generated using the same A* pathfinding algorithm as in BiBo, while interaction point generation follows the procedure used in UniHSI. The task execution process is illustrated in Fig.~\ref{fig:fig13}
\subsubsection{CLoSD}
CLoSD~\citep{tevet2024closd} use a motion diffusion to drive a tracking policy~\citep{luo2023perpetual} to perform human-scene interaction in physical environment. It is controlled by a rule-based finite state machine (FSM). The FSM determines the next action based on the current state and a programmatic script, outputting the subsequent 60-frame action command that includes the motion caption, target positions of the feet, hands, and head, as well as the standing position and facing direction.

We employ the official implementation\footnote{https://github.com/GuyTevet/CLoSD} and the {\small\itshape Multitask} model checkpoint. The state sequence is generated according to the scene objects and task. Specifically, for each interaction task, we define an update function and a transition rule. The update function outputs the next action command and checks whether a transition should occur. When all interactions in the current group meet their transition conditions, the FSM proceeds to the next group of interactions (where interactions within the same group occur simultaneously). The transition rules follow the same criteria of task success, as described in Sec.~\ref{sec:task_completion}.

In the {\small\itshape reach} state update function, the A* algorithm is first applied to perform pathfinding. If the agent is oriented toward the next waypoint within a 45° cone, the target position is set to that waypoint (clipped to a maximum distance of 1.2m), and the motion caption prompt is {\small\itshape ``a person is walking.''} Otherwise, the facing direction is adjusted toward the next waypoint, and the prompt is set to {\small\itshape ``a person is turning around in place.''} The configurations for {\small\itshape sit}, {\small\itshape sleep}, and {\small\itshape touch} actions follow those defined in UniHSI. For the {\small\itshape lift} action, the target positions of both hands are initially placed 0.3m to the left and right sides of the object, then directed toward the object’s center to facilitate grasping. Finally, the pelvis height is set to 0.86m (the normal standing height) to guide the agent to stand upright. A {\small\itshape reach} state is inserted between consecutive interactions to enable navigation across interaction points.

\subsubsection{Ablation Study}
\label{sec:com:t2m:abl}
In the ablation study presented in Sec.~\ref{sec:task_completion}, we analyze the impact of several components in both the compiler and executor. In the compiler, we ablate the voting mechanism and the use of image labels, while in the executor, we ablate the use of actual executed motion input, previously generated motion input, and IK post-optimization.

To ablate the voting mechanism, we use a single VLM to analyze motion attributes instead of aggregating outputs from multiple VLM instances. The image label is ablated by prompting the VLM to directly predict image coordinates and facing angles. For the executor, we conduct ablations by extending future motion using only the previously generated motion or the actual executed motion. Specifically, the same motion sequence is provided as input to both the Diffusion Model and the VAE during inference. Finally, we ablate IK post-optimization by directly using the raw generated motion without applying inverse kinematics refinement.

\subsection{Motion Quality}
\subsubsection{MotionLCM}
MotionLCM~\citep{dai2024motionlcm} adopts a Transformer-based Latent Diffusion architecture, incorporates Consistency training to enable few-step generation, and integrates control signals through a ControlNet module.

We use MotionLCM v2, adopting the official implementation\footnote{https://github.com/Dai-Wenxun/MotionLCM}
 and following the provided training scripts and configurations to train the VAE, Consistency Model, and ControlNet. In addition to the original TM2T metrics, we introduce two additional evaluation pipelines to assess Physical Plausibility and Control Accuracy.

The Physical Plausibility evaluation follows the implementation provided in CLoSD, whereas the Control Accuracy evaluation is performed under the same experimental setting as BiBo. Specifically, a single joint is randomly selected from the head, hand, or foot, and the ground-truth joint position at frame 40 is used as the control signal. The mean absolute error (MAE) between the generated joint position and the control signal is then calculated to quantify the control error.

Training and evaluation is conducted on a workstation with 8× RTX 4090 GPUs, 256 GB DDR4 RAM, and an Intel(R) Xeon(R) Gold 6226R CPU @ 2.90GHz, running Ubuntu 24.04 with CUDA 11.8 and PyTorch 2.3.

\subsubsection{MotionStreamer}
MotionStreamer~\citep{xiao2025motionstreamer} adopts a latent diffusion architecture, where a Transformer-based diffusion model performs next-token prediction, and a CNN-based VAE decoder reconstructs the motion from tokens, enabling the generation of high-fidelity motions of arbitrary length.

We use the official implementation\footnote{https://github.com/zju3dv/MotionStreamer}. The original version employs a 272-dimensional motion representation\footnote{https://github.com/Li-xingXiao/272-dim-Motion-Representation}, which is incompatible with our evaluation pipeline. Therefore, we modify it to use the 263-dimensional motion representation in HumanML3D, and train the model following the provided scripts and configurations. The generated motions are exported into the evaluation pipeline to assess motion quality and physical plausibility.

Training is conducted on a workstation equipped with 8× RTX A6000 GPUs, 256 GB DDR4 RAM, and an Intel(R) Xeon(R) Gold 6226R CPU @ 2.90 GHz, running Ubuntu 24.04 with CUDA 11.8 and PyTorch 2.3. Evaluation is performed on the same machine as MotionLCM.

\subsubsection{MoGenTS}
MoGenTS~\citep{yuan2024mogents} is based on a masked modeling framework, consisting of a Motion Transformer and a CNN-based vector quantization VAE (VQ-VAE). It introduces a masking mechanism that leverages both spatial skeletal relationships and temporal dependencies, enabling high-quality motion generation. We use the official implementation\footnote{https://github.com/weihaosky/mogents} and the released checkpoints. The generated motions are then exported to the evaluation pipeline to assess motion quality and physical plausibility. The evaluation is conducted on the same machine as MotionLCM.

\subsubsection{MoConVQ}
MoConVQ~\citep{yao2024moconvq} is an end-to-end physical motion generator built upon a VQ-VAE architecture, which operates within its own simulation environment. We employ the official implementation\footnote{https://github.com/heyuanYao-pku/MoConVQ} and released checkpoint for all experiments. For quantitative evaluation, we directly adopt the results reported in CLoSD to ensure consistency and comparability across methods. In the qualitative experiments illustrated in Fig.~\ref{fig:fig5}, we evaluate MoConVQ under two representative prompts: (d) {\small\itshape ``A person is boxing with someone, and kicking at the air.''} and (e) {\small\itshape ``A person is sitting on a sofa and waving a hand above the head.''} Since the simulation environment of MoConVQ does not natively support static mesh colliders like sofa, we follow the approach in this repository\footnote{https://github.com/heyuanYao-pku/Control-VAE}
, and use a rectangular block with the same height as the sofa cushion and very large mass as a substitute. The evaluation is conducted on the same machine as MotionLCM.

\subsubsection{Dip \& CLoSD}
DiP~\citep{tevet2024closd} is a real-time motion diffusion model capable of generating motions of arbitrary length, built upon a Transformer architecture. CLoSD employs a tracking policy to follow the motions generated by DiP in physical environment. DiP extends future motion from the actual executed motion, thereby adapting physical feedback.

We use the official implementation\footnote{https://github.com/GuyTevet/CLoSD}. In the text-to-motion experiments in Tab.~\ref{tab:text2motion}, we adopt the released {\small\itshape t2m} checkpoint, while in the control accuracy experiments in Tab.~\ref{tab:con}, we use the {\small\itshape multi-target} checkpoint and adopt the same setting as other comparison methods. The motions generated by DiP and CLoSD are exported using the provided scripts and evaluated through a unified evaluation pipeline as other comparison methods. The evaluation is conducted on the same machine as MotionLCM, using IsaacGym Preview 4.

\subsubsection{Ablation Study}
In Tab.~\ref{tab:con},~\ref{tab:jerk}, and~\ref{tab:text2motion_ablation}, we ablate the LDM by directly training the diffusion model on raw motion sequences, and ablate Causal Attention by removing the attention mask during both training and inference. The inputs of actual executed motion and previously generated motion are further ablated by using only one of them during inference, following the same setting described in Sec.~\ref{sec:com:t2m:abl}. 

%% file: sections/F_sup_experiment.tex
\section{Supplementary Experiments}
\subsection{Rendering}
In the experiments, we use Blender Python API 4.2.0 EEVEE Next pipeline to render the scene images during the planning process. 

For visualization, we adopt three approaches:
\begin{enumerate}
    \item Headless server environments: We perform visualization rendering in Blender using the EEVEE Next pipeline, where humanoid body are represented as skeletal lines.
    \item Debugging environments and Fig.~\ref{fig:fig16},~\ref{fig:fig17} and~\ref{fig:fig18}: We use the built-in visualization window provided by IsaacGym for rendering.
    \item Case study in Fig.~\ref{fig:fig5} and video demonstrations: We conduct simulations in IsaacGym, export the root states and meshes of all objects, and render them in Unity. The humanoid body is reconstructed in Unity\footnote{https://unity.com/} based on IsaacGym XML-based humanoid body file.
\end{enumerate}

\begin{table}
\centering
\scriptsize
\setlength{\tabcolsep}{5pt} 
\begin{minipage}{0.64\linewidth}
\centering
\caption{Comparison between different VAEs on motion reconstruction error, using HumanML3D. \textbf{Bold} and \underline{underline} is the best and second best, respectively.}
\begin{tabular}{l|cc|cccc}
\hline
\textbf{Method} & \textbf{MotionLCM} & \textbf{MoGenTS} & \textbf{VAE} & \textbf{w/ Causal}& \textbf{w/ AE} & \textbf{w/ VQ} \\
\hline
FID $\downarrow$   &0.025&\textbf{0.006}& 0.021 &\underline{0.012}  & 0.031 & 0.107 \\
MPJPE $\downarrow$ &27.44&\underline{16.35}& 20.36 &\textbf{7.58}   & 31.72 & 39.82 \\
\hline
\end{tabular}
\label{tab:vae}
\end{minipage}
\hfill
\begin{minipage}{0.32\linewidth}
\centering
\caption{Impact of latent dimension of the Causal VAE on motion reconstruction error.}
\begin{tabular}{l|ccc}
\hline
\textbf{Latent Dim} & \textbf{32} & \textbf{64} & \textbf{128}\\
\hline
FID  $\downarrow$ & 0.015& \underline{0.012}  &\textbf{0.010} \\
MPJPE $\downarrow$ & 8.04& \underline{7.58}   & \textbf{6.76} \\
\hline
\end{tabular}
\label{tab:vae_dim}
\end{minipage}
\end{table}

\subsection{Variants of VAE}
\textbf{Setting.} We experimented with several types of VAEs to convert motion sequences into latent tokens for higher reconstruction quality. Specifically, we tested:
\begin{enumerate}
    \item The original VAE architecture with a Skip-Transformer encoder–decoder structure.
    \item Adding causal attention, as described in Sec.~\ref{sec:met:mde}.
    \item Introducing causal autoregressive next-token prediction decoding — specifically, like the behavior of GPT. Instead of feeding $N$ zero tokens at once and decoding all $N$ motion frames simultaneously, we iteratively append one zero token at a time and decode $N$ frames across $N$ steps. All other hyperparameters remain identical with the original setting.
    \item Incorporating vector-quantized VAE (VQ-VAE), implemented using an residual-quantization VAE (RQ-VAE) with 1,024 code entries, 6 residual quantizers, and an exponential moving average (EMA) decay factor of $\mu$ = 0.99.
    
\end{enumerate}

We train these VAE models on the train split of HumanML3D, and evaluate FID and Mean Per Joint Position Error (MPJPE) on the test split. FID evaluates the similarity between the distribution of the original and reconstructed dataset, and MPJPE evaluates the reconstruction error.

\textbf{Result.} According to the experimental results in Tab~\ref{tab:vae}, the (2) Causal Attention configuration achieves the best performance, with a MPJPE of $7.58$mm.
We further test models with different latent dimensions under the Causal Attention setting, and the results in Tab.~\ref{tab:vae_dim} show that larger latent dimensions further improve performance.
To balance computational efficiency and reconstruction accuracy, we choose a latent dimension of 64.

\begin{table}
\centering
\caption{Comparison between different VLM variants on task success rate in random generated scene. \textbf{Bold} and \underline{underline} represent the best and second best performance, respectively. \textcolor{gray}{Gray} color denotes using ground-truth action plan, which is excluded in the comparison.}
\scriptsize
\resizebox{1.0\textwidth}{!}{
\begin{tabular}{l|cccccc|ccc}
\hline
\multirow{2}{*}{\textbf{Method (\%)}} & \multicolumn{6}{c|}{\textbf{Single Interaction}} & \multicolumn{3}{c}{\textbf{Composite Task}} \\
\cline{2-7} \cline{8-10}
 & Reach $\uparrow$ & Watch $\uparrow$ & Sit $\uparrow$ & Sleep $\uparrow$ & Touch $\uparrow$ & Lift $\uparrow$ & Simple $\uparrow$& Medium $\uparrow$ & Hard $\uparrow$ \\
\hline
BiBo (ours, w/ Sonnet 3.5, $\sim175B$) & \textbf{99.18} & \underline{98.31} & \underline{87.54} & \underline{91.33} & \underline{70.57} & \underline{59.52} & \underline{50.00} & \underline{34.62} & \underline{19.05}\\
BiBo (ours, w/ Qwen2.5-VL, $\sim72$B) & \underline{98.09} & 98.87 & 72.73 & 78.67 & 68.44 & 55.10 & 25.00 & 11.54 & 14.29\\
BiBo (ours, w/ GPT-4o mini, $\sim8$B) & 91.28 & 92.66 & 48.15 & 16.00 & 56.74 & 52.38 & 19.12 & 13.46 & 9.52\\

\hline

BiBo (ours, w/ GPT-4o, $\sim200$B)         & \textbf{99.18} & \textbf{99.62} & \textbf{95.84} & \textbf{94.89} & \textbf{86.05} & \textbf{65.42} & \textbf{58.82} & \textbf{36.54} & \textbf{27.78} \\
\textcolor{gray}{BiBo (ours, w/ GT plan)} & \textcolor{gray}{98.91} & \textcolor{gray}{99.06} & \textcolor{gray}{96.75} & \textcolor{gray}{93.33} & \textcolor{gray}{87.23} & \textcolor{gray}{70.41} & \textcolor{gray}{61.76} & \textcolor{gray}{44.23} & \textcolor{gray}{42.86} \\
\hline
\end{tabular}
}
\label{tab:vlm}
\end{table}

\subsection{Variants of Off-the-shelf VLMs}
\textbf{Setting.} We evaluate BiBo’s scaling capability across VLMs of different sizes, and assess the generalizability of prompt design. Specifically, we use the large-scale Claude Sonnet 3.5 ($\sim175$B), the medium-scale Qwen2.5-VL ($\sim72$B), and the small-scale GPT-4o mini ($\sim8$B). The parameter size of Qwen2.5-VL is publicly available \citep{bai2025qwen2}, while the sizes of Claude Sonnet 3.5 and GPT-4o mini are referenced from \citet{abacha2024medec}. In implementation, we directly replace the API endpoint of GPT-4o with that of each comparison VLMs. Each task is evaluated once under a randomly initialized pose.

\textbf{Result.} The results in Tab.~\ref{tab:vlm} show that when using the prompt originally designed for GPT-4o, BiBo achieves a comparable task success rate on the similarly scaled Claude Sonnet 3.5, demonstrating generalization capability. As the size of the VLM increases, the task success rate improves, reflecting scaling ability.

\begin{mytbox}[label={box:vg_json}]{Prompt Example for Visual Grounding (JSON Coordinates)}
\textbf{SYSTEM\textcolor{red}{*}:} You are a precise visual grounding model that outputs JSON coordinates only.
\\
\\
\textbf{USER\textcolor{red}{*}:} You are given an image. Your task is to locate the \underline{\textless Object Category\textgreater} in the image. 
Please output the approximate \textbf{center coordinates} of this object.
\\
\\
The coordinate system follows the image pixel convention:

- The origin (0, 0) is at the top-left corner.

- The x-axis increases to the right.

- The y-axis increases downward.

Return coordinates in pixels relative to the image resolution.
\\
\\
Return only a JSON object, for example:
\begin{verbatim}
{"x": 180, "y": 240}
\end{verbatim}
\end{mytbox}

\begin{mytbox}[label={box:vg_label}]{Prompt Example for Visual Grounding (Numerical Labels)}
\textbf{SYSTEM\textcolor{red}{*}:} You are a precise visual grounding model that identifies numerical labels in an image and outputs only structured results.
\\
\\
\textbf{USER\textcolor{red}{*}:} You are given an image that contains 64 numerical labels (each labeled with a number from 1 to 64).
\\
\\
Your task:

- Select the label that lies on the visible surface of the \underline{\textless Object Category\textgreater}.

- Avoid choosing any number that lies on occluding or overlapping objects that block the target.

- Then, output only one label number (an integer between 1 and 64) that best fits the requirement.
\\
\\
Output format: 

- Wrap your final answer between $>>>$ and $<<<$ for easy parsing.
\end{mytbox}

\subsection{Accuracy of the Three-Stage VQA}
\textbf{Setting.} For basic motion attribute analysis, we incorporate a voting mechanism to enhance robustness. In the agent pose reasoning process, a series of textual label descriptions (e.g., object orientations predicted by Orient Anything) are introduced to improve the VLM’s spatial understanding. During key joint position generation, we overlay a grid of numerical labels on the image to facilitate visual grounding. 

We conduct ablation studies on these components to evaluate their individual contributions. The voting mechanism is ablated by using only a single VLM instance for analysis, and the textual label descriptions are ablated by using only the raw image labels without any additional contextual information in the prompt. The effect of the label grid is examined by comparing the joint generation accuracy with and without the grid, using the prompts shown in Box~\ref{box:vg_json} and Box~\ref{box:vg_label}, respectively.

The experimental designs are specified as follows:

\begin{enumerate}
    \item \textbf{Basic Attribute Analysis.} We randomly sample 100 test cases during the task evaluation process and manually annotate the corresponding ground-truth plans. Each test case includes a motion caption, an agent state, a scene observation, and a ground-truth plan. The VLM is prompted as described in Box~\ref{box:basic_analysis1} and Box~\ref{box:basic_analysis2}. The output of the VLM is considered correct if it matches the ground-truth annotation for each motion attribute.
    
    \item \textbf{Agent Pose Reasoning.} Following the same evaluation process as in (1), we select 100 interaction cases involving the agent’s position and facing direction. For the \emph{watch} task, we exclude the evaluation of agent–anchor distance since multiple plausible spatial configurations may exist.
    
    \item \textbf{Key Joint Generation.} This is formulated as a visual grounding task. We construct test cases using the COCO 2017 dataset~\citep{lin2014coco}\footnote{https://cocodataset.org/} and evaluate whether the coordinates predicted by the VLM fall within the ground-truth instance boundaries, allowing a tolerance of 10 pixels. Only images that contain a category with exactly one instance are used, where the instance area should occupy 4\%–25\% of the entire image. Based on these criteria, we filter 500 samples from the 5,000-image validation split and perform the evaluation using GPT-4o mini for cost efficiency.
\end{enumerate}

\begin{figure}[t!]
    \centering
    \includegraphics[width=1.0\textwidth]{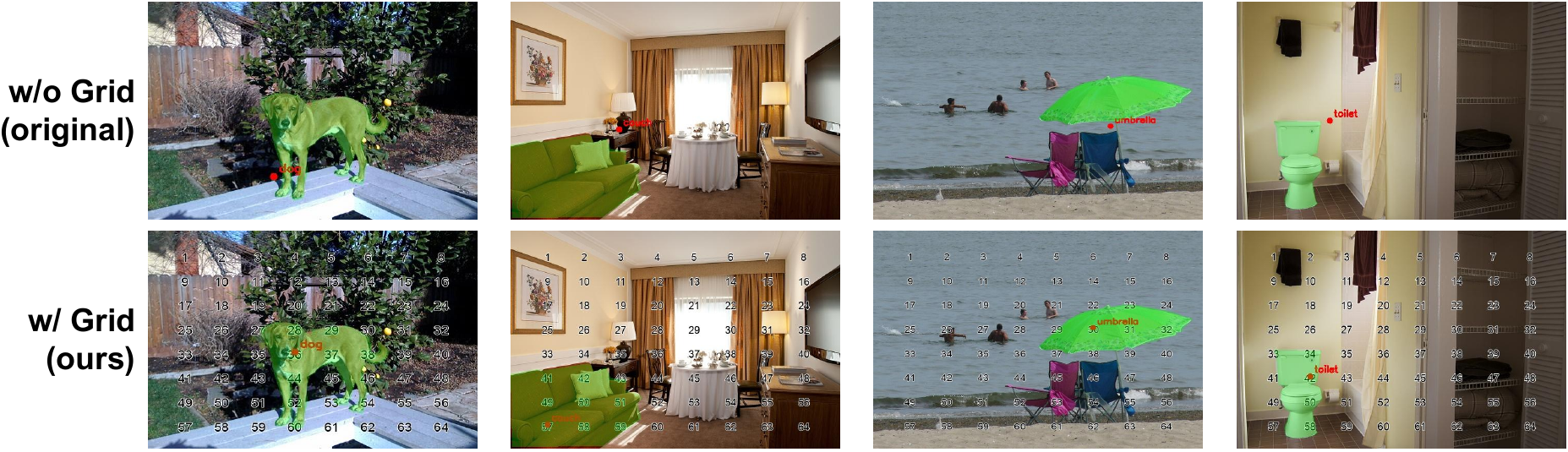}
    \caption{Impact of the label grid on visual grounding of VLM. The result shows that label grid effectively enhance the grounding accuracy.}
    \label{fig:fig14} 
\end{figure}

\textbf{Result.} According to the experimental results in Tab.~\ref{tab:abl_ana},~\ref{tab:abl_pos} and~\ref{tab:abl_joi}, the voting mechanism achieved a relative improvement of 10.3\% in basic attribute analysis accuracy. The combination of Orient Anything and the rule-based textual label description yielded a 31.9\% relative improvement in pose reasoning accuracy—an enhancement primarily observed in the model’s understanding of object orientation. Meanwhile, the label grid led to a 55.9\% relative improvement in localization performance, which aligns with the conclusions of \citet{yang2023set} and \citet{chae2024grid}. The visualization of the grounding results are shown in Fig.~\ref{fig:fig14}.

\begin{table}[t]
\centering
\setlength{\tabcolsep}{4pt}
\renewcommand{\arraystretch}{1.2}
\scriptsize
\begin{minipage}[t]{0.3\textwidth}
\centering

\caption{Impact of voting mechanism on motion attribute analyzing accuracy.}
\begin{tabular}{lc}
\toprule
\textbf{Method} & \textbf{Accuracy$\uparrow$} \\
\midrule
Analyzing (w/o Voting) & 78\% \\
Analyzing  & \textbf{86\%} \\
\bottomrule
\end{tabular}
\label{tab:abl_ana}
\end{minipage}
\hfill
\begin{minipage}[t]{0.3\textwidth}
\centering

\caption{Impact of label description on agent pose reasoning accuracy.}
\begin{tabular}{lc}
\toprule
\textbf{Method} & \textbf{Accuracy$\uparrow$} \\
\midrule
Pose (w/o Description) & 69\% \\
Pose  & \textbf{91\%} \\
\bottomrule
\end{tabular}
\label{tab:abl_pos}
\end{minipage}
\hfill
\begin{minipage}[t]{0.3\textwidth}
\centering

\caption{Impact of label grid on key joint position generation accuracy.}
\begin{tabular}{lc}
\toprule
\textbf{Method} & \textbf{Accuracy$\uparrow$} \\
\midrule
Joint (w/o Label Grid) & 42.2\% \\
Joint  & \textbf{65.8\%} \\
\bottomrule
\end{tabular}
\label{tab:abl_joi}
\end{minipage}

\end{table}

\begin{table}[t]
\centering
\caption{The impact of visual information, reflection and speed control on task success rate. \textbf{Bold} and \underline{underline} represent the best and second best performance.}
\scriptsize
\resizebox{1.0\textwidth}{!}{
\begin{tabular}{l|cccccc|ccc}
\hline
\multirow{2}{*}{\textbf{Method (\%)}} & \multicolumn{6}{c|}{\textbf{Single Interaction}} & \multicolumn{3}{c}{\textbf{Composite Task}} \\
\cline{2-7} \cline{8-10}
 & Reach $\uparrow$ & Watch $\uparrow$ & Sit $\uparrow$ & Sleep $\uparrow$ & Touch $\uparrow$ & Lift $\uparrow$ & Simple $\uparrow$& Medium $\uparrow$ & Hard $\uparrow$ \\
\hline
BiBo (ours, w/o Visual Info.) & 96.19 & 89.83 & 84.18 & \underline{90.67} & 76.24 & \underline{62.93} & \underline{54.41} & 30.77 & 21.43\\
BiBo (ours, w/o Reflection) & \underline{96.46} & \underline{95.48} & \underline{86.87} & 81.33 & \underline{81.56} & 61.56 & 50.00 & \underline{32.69} & \underline{23.81}\\
BiBo (ours, w/o Speed Control) & 73.57 & 84.18 & 72.73 & 69.33 & 51.77 & 27.89 & 38.24 & 15.38 & 9.52\\
\hline
BiBo (ours)         & \textbf{99.18} & \textbf{99.62} & \textbf{95.84} & \textbf{94.89} & \textbf{86.05} & \textbf{65.42} & \textbf{58.82} & \textbf{36.54} & \textbf{27.78} \\
\hline
\end{tabular}
}
\label{tab:sup_abl}
\end{table}

\begin{table}[t]
\centering
\caption{The control error (MAE) of BiBo on HumanML3D dataset, including facing direction (Rot.), moving speed (Vel.) and joint positions.}
\setlength{\tabcolsep}{8pt}
\renewcommand{\arraystretch}{1.2}
\scriptsize
\begin{tabular}{l|c|c|cccccc}
\hline
\multirow{2}{*}{\textbf{Cond}} & 
\multirow{2}{*}{\textbf{Rot. (rad)}} & 
\multirow{2}{*}{\textbf{Vel. (m/s)}} & 
\multicolumn{6}{c}{\textbf{Joint Position (m)}} \\
\cline{4-9}
 &  &  & Pelvis & Head & L Hand & R Hand & L Foot & R Foot \\
\hline
MAE $\downarrow$ & 0.177207 & 0.015499 & 0.024649 & 0.033833 & 0.054909 & 0.057240 & 0.032403 & 0.032687 \\
\hline
\end{tabular}
\label{tab:test_mae}
\end{table}

\subsection{Supplementary Ablation Study}

\textbf{Setting.} For the compiler, we employ a VLM for planning, thereby incorporating visual information. During the reasoning process, a rule-based reflection mechanism (refer to Sec.~\ref{sec:com:pln:ref}) is integrated for error detection and self-correction. For the executor, an additional moving-speed control signal is introduced. We evaluate the contributions of these components through ablation and test the task success rate after each ablation. Each task runs once in the ablation groups.

The visual information ablation is performed by removing all image inputs in the first two VQA stages while retaining the label descriptions (e.g., the relative distance and orientation of each label with respect to the anchor). Since key joint generation is based on image, we keep visual information in the last VQA stage. To ablate the reflection mechanism, all rule-based detections are removed. For the moving-speed ablation, the corresponding condition mask in the motion diffusion is set to zero, and all speed-related descriptions (e.g., walking slowly, turning slowly, refer to Sec.~\ref{sec:com:state:navi}) are removed from the compiler.

\begin{figure}[t!]
    \centering
    \includegraphics[width=1.0\textwidth]{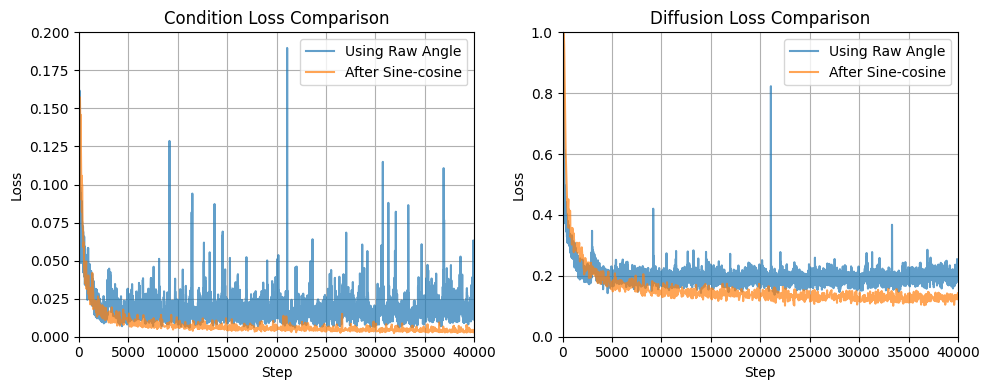}
    \caption{Visualization of the training loss variations with and without using sine–cosine encoding of the facing direction. For both the Diffusion Loss and Condition Loss, the sine–cosine encoding yields more stable training and faster convergence.}
    \label{fig:fig15} 
\end{figure}

\textbf{Result.} The results in Tab.~\ref{tab:sup_abl} show that visual information and reflection both contribute to improving the planning quality of the compiler, while the moving speed control effectively enhances locomotion in the scene.

\subsection{Control Accuracy}
\textbf{Setting.} First, we evaluate the control accuracy of BiBo on HumanML3D. The control conditions include the facing direction measured in $rad$, the moving speed measured in $m/s$, and the joint positions measured in $m$. The control error is represented using the Mean Absolute Error (MAE).
Second, a sine-cosine encoding is adopted for the facing direction during training. We compare the training convergence curves of raw facing angle and the sine-cosine representation.

\textbf{Result.}  The results in Tab.~\ref{tab:test_mae} demonstrate that the control parameters provide effective spatial guidance for motion generation.
As shown in Fig.~\ref{fig:fig15}, employing sine–cosine facing direction encoding improves both the training stability and convergence.

\subsection{Hyperparameter Selection}

\begin{table}[t]
\centering
\setlength{\tabcolsep}{8pt}
\renewcommand{\arraystretch}{1.2}
\scriptsize
\caption{Impact of hyperparameter combinations on generation efficiency and motion quality, using HumanML3D. \textbf{Bold} and \underline{underline} denotes the best and second best performance.}
\begin{tabular}{c | c | c | c | c c | c c c c}
\hline
\multirow{2}{*}{\textbf{Hid. Dim.}} &
\multirow{2}{*}{\textbf{F.F. Dim.}} &
\multirow{2}{*}{\textbf{ Head }} &
\multirow{2}{*}{\textbf{LR}} &
\multicolumn{2}{c|}{\textbf{Efficiency}} &
\multicolumn{4}{c}{\textbf{Motion Quality}} \\
\cline{5-10}
 & & & & \textbf{Params (M) $\downarrow$} & \textbf{AITS$\downarrow$} & \textbf{FID$\downarrow$} & \textbf{R.P.1$\uparrow$} & \textbf{R.P.2$\uparrow$} & \textbf{R.P.3$\uparrow$} \\
\hline
256 & 1024 & 4 & 5e-4 & \underline{28.541} & \underline{0.047} & \textbf{0.076} & \underline{0.542} & \textbf{0.738} & \textbf{0.829}\\
256 & 1024 & 4 & 1e-3 & \underline{28.541} & \underline{0.047} & 0.094 & 0.536 & 0.729 & 0.822\\
256 & 1024 & 4 & 2e-3 & \underline{28.541} & \underline{0.047} & 0.091 & \textbf{0.543} & \underline{0.735} & 0.826 \\
128 & 512 & 4 & 5e-4 & \textbf{20.584} & \textbf{0.044} & 0.127 & 0.505 & 0.703 & 0.804 \\
512 & 2048 & 8 & 5e-4 & 59.495 & 0.061 & \underline{0.085} & 0.534 & 0.732 & \underline{0.827}\\
\hline
\end{tabular}
\label{tab:hyperparam_selection}
\end{table}

\textbf{Setting.} We investigate the effects of different learning strategies and model sizes on motion generation.
For the learning rate, we experiment with 5e-4, 1e-3, and 2e-3.
For the model size, we select hidden dimensions of 128, 256, and 512, with the feed-forward layer dimension scaled proportionally.
For the 512-dimensional model, we use 8 attention heads.
All other parameters remain consistent with the current configuration.

We evaluate both computational efficiency and generation quality. Computational efficiency is measured by the number of parameters and AITS, while generation quality is assessed using FID and Top-$1\sim3$ R-Precision.

\textbf{Result.} As shown in Tab.~\ref{tab:hyperparam_selection}, the current configuration (256 hidden dimension, 1024 feed-forward dimension, 4 attention heads, and a 5e-4 learning rate) achieves the best generation quality, while maintaining computational efficiency comparable to smaller models.
Due to the limited dataset size, larger models significantly increase computational cost without yielding better performance.

\begin{table}[t]
\centering
\setlength{\tabcolsep}{8pt}
\renewcommand{\arraystretch}{1.2}
\scriptsize
\caption{Comparison of different prefix strategies under two preset conditions.}
\begin{tabular}{l | cccc | cccc}
\hline
\multirow{2}{*}{\textbf{Method}} &
\multicolumn{4}{c|}{\textbf{Ground-truth Motion Preset}} &
\multicolumn{4}{c}{\textbf{Prefix Motion Preset}} \\
\cline{2-9}
 & \textbf{FID↓} & \textbf{R.P.1↑} & \textbf{R.P.2↑} & \textbf{R.P.3↑} &
   \textbf{FID↓} & \textbf{R.P.1↑} & \textbf{R.P.2↑} & \textbf{R.P.3↑} \\
\hline
No Prefix     & \textbf{0.068} & \textbf{0.552} & \textbf{0.747} & \textbf{0.834} & 0.156 & 0.492 &0.682 & 0.786 \\
\hline
All-zero Prefix          & 0.086 & 0.511 & 0.713 & 0.809 & \underline{0.106} & 0.501 & 0.703 & 0.799 \\
Learnable Prefix Token   & 0.123 & 0.523 & 0.721 & 0.818 & 0.148 & \underline{0.504} & \underline{0.704} & \underline{0.803} \\
Stance Motion Prefix     & \underline{0.076} & \underline{0.542} & \underline{0.738} & \underline{0.829} & \textbf{0.087} & \textbf{0.525} & \textbf{0.721} & \textbf{0.816} \\
\hline
\end{tabular}
\label{tab:prefix_comparison}
\end{table}

\subsection{Motion Prefix}

\textbf{Setting.} We prepend a prefix to all motion sequences in the dataset.
This strategy improves data utilization efficiency and enhances the model’s ability to initiate motions from scratch or dynamically transfer to new motions.
We explore three different prefix augmentation strategies:
\begin{enumerate}
    \item Adding an all-zero prefix.

    \item Extracting a stance motion frame from the dataset and replicating it for 20 frames.

    \item Introducing a learnable prefix token.
\end{enumerate}

We evaluate the generation quality on the HumanML3D dataset. During testing, we preset an initial motion segment and let the model to iteratively extend the future motion, after which the preset portion is trimmed off to compute FID and R-Precision.
Two preset motion conditions are tested:
(1) using the first 20 frames of ground-truth motion (as in \citet{tevet2024closd}), and
(2) using the newly added prefix.

\textbf{Result.} Experimental results in Tab.~\ref{tab:prefix_comparison} show that introducing a prefix enhances the model’s ability to generate motion from scratch, rather than relying on a preset motion history.
Among the tested strategies, the stance-motion prefix yields the best performance.

\subsection{Case Study}

\begin{figure}[t!]
    \centering
    \includegraphics[width=1.0\textwidth]{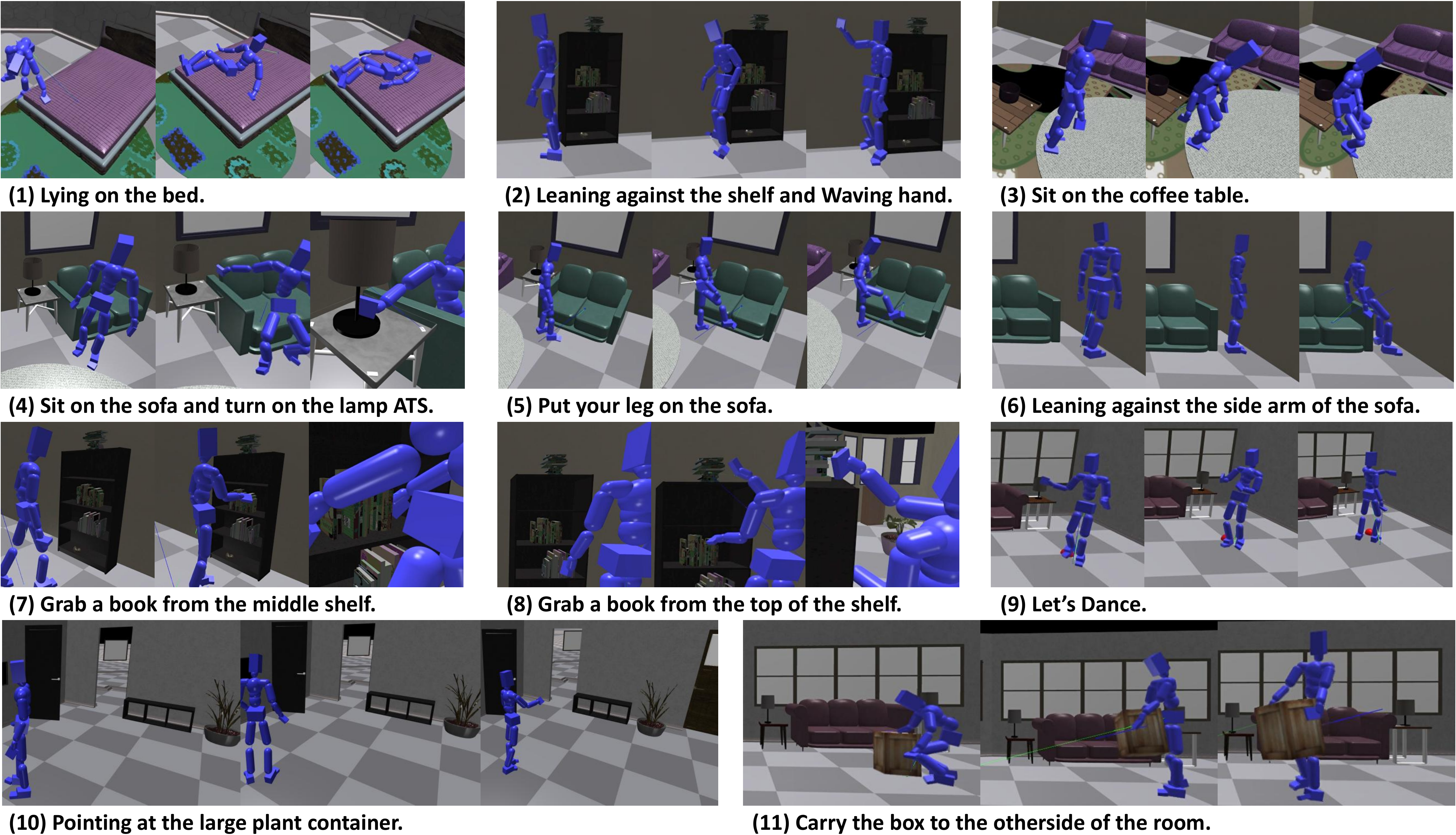}
    \caption{Visualization of BiBo in IsaacGym. BiBo can perform one type of interaction across multiple object categories (in 3, 4), and execute multiple types of interactions on the same object (in 4, 5, 6).
It can generate diverse motions conditioned on text (in 9), and supports precise control (in 7, 8), composite action (in 2, 4), long-range interactions (in 10) and manipulation (in 11).}
    \label{fig:fig16} 
\end{figure}

\textbf{Basic Interactions.} We evaluate BiBo’s ability to perform various types of interactions, including basic interactions (e.g., sit, sleep), text-driven motion generation (e.g., dance), long-range interactions (e.g., pointing at), composite actions, and manipulation tasks (e.g., transport).

The prompts used for testing and the corresponding execution results in the simulator are shown in Fig.~\ref{fig:fig16}.
According to the results, BiBo demonstrates diversity in interaction targets (in 3, 4), interaction modes (in 4, 5, 6), and control modalities (in 9).
It also supports precise control (in 7, 8), composite actions (in 2, 4), long-range interactions (in 10), and manipulation tasks (in 11).

For more visual demonstration, please visit our Hugging Face repository~\footnote{https://huggingface.co/Behavia/BEHAVIA}.

\begin{figure}[t!]
    \centering
    \includegraphics[width=1.0\textwidth]{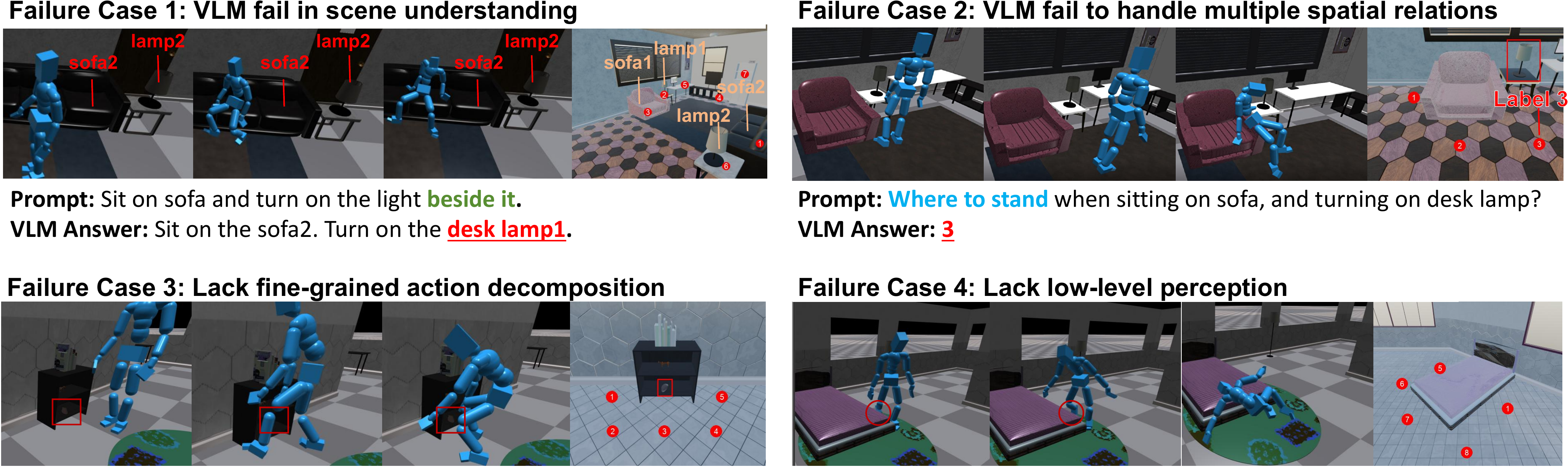}
    \caption{Visualization of failure cases of BiBo. Case 1 and 2 are related to the limitation of VLM, and Case 3 and 4 are related to the lack of explicit scene geometry modeling (e.g. scene voxel, point cloud) in executor.}
    \label{fig:fig17} 
\end{figure}

\textbf{Failure Cases.} We visualize representative failure cases during task execution. The results are shown in Fig.~\ref{fig:fig17}.
In case (1), the VLM misinterprets the scene layout (or mismatches textual labels and scene objects), leading to the selection of an incorrect interaction target.
In case (2), the VLM focuses only on the agent’s position relative to the desk lamp while ignoring the sofa’s orientation.
In cases (3) and (4), because the executor passively receives physical feedback rather than actively parsing the geometric structure of the scene, it shows limitations in fine-grained action decomposition and proactive obstacle avoidance.

Our method still has limitations, but we believe that as the multimodal capabilities of off-the-shelf VLMs and 3D encoding technologies continue to advance, these problems are expected to be resolved in the near future.

\begin{figure}[t!]
    \centering
    \includegraphics[width=1.0\textwidth]{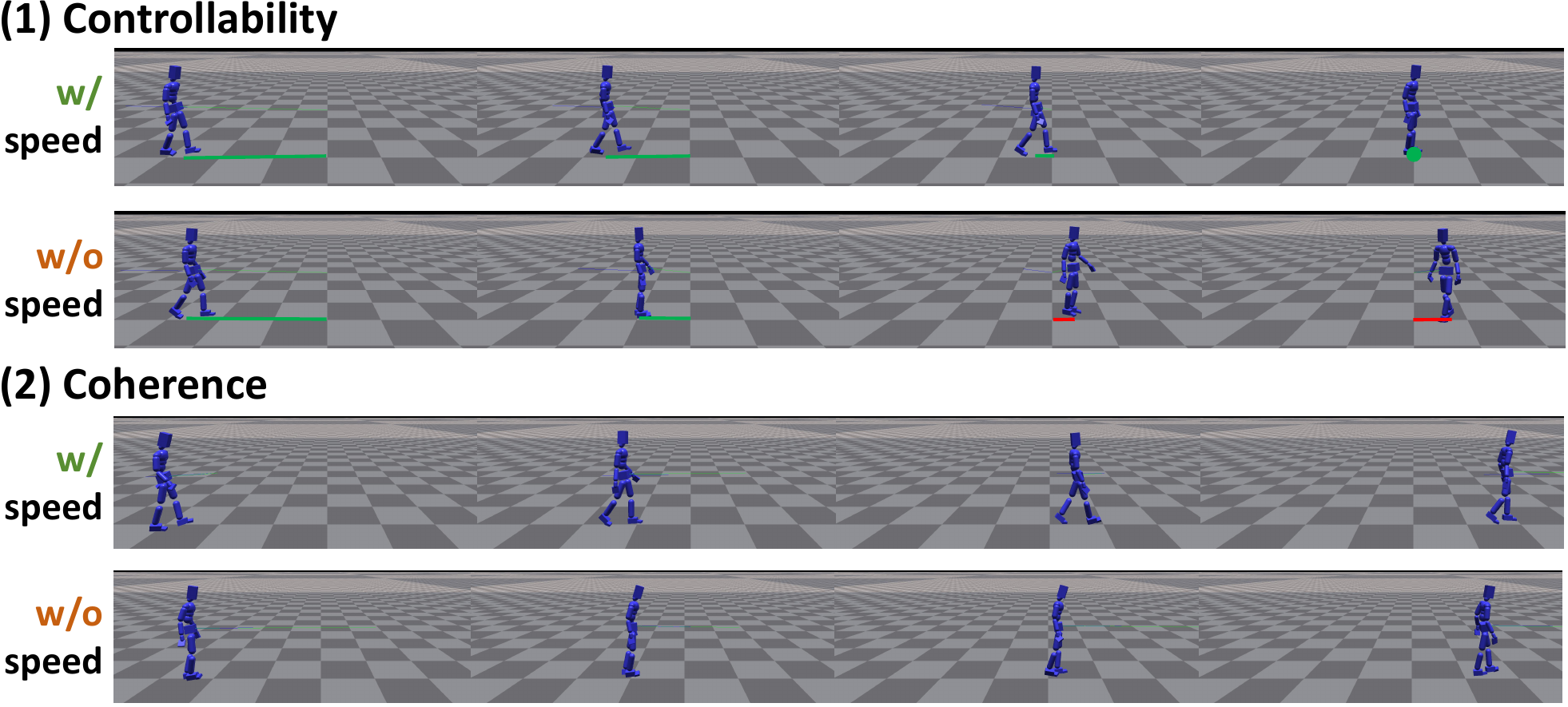}
    \caption{Visualization of locomotion with and without moving-speed control.
In (1), the agent with speed control precisely reaches the target location, enhancing controllability.
In (2), the agent without speed control tends to stand still between successive walking motions, as it cannot determine whether to stop or keep moving, lacking coherence.}
    \label{fig:fig18} 
\end{figure}

\textbf{Ablation of Speed Control.} During experiment, we observe that the moving speed not only enhances controllability, but also serves as a signal indicating whether to continue the motion or to stop the current one, which enhance motion coherence.

As shown in Fig.~\ref{fig:fig18}, in (1), the agent with speed control precisely reaches the target location, while the agent without speed control overshoots the target and then turns back.
In (2), the agent with speed control keeps walking status between successive walking motions, while the agent without speed control tends to pause between successive walking motions, as it cannot determine whether to stop or keep moving. The agent with speed control demonstrates better controllability and motion coherence.
